\documentclass[sigconf]{acmart}

\usepackage[utf8]{inputenc}
\graphicspath{ {images/} }
\usepackage{geometry}
\usepackage{balance}
\usepackage{amsmath}
\usepackage{adjustbox}
\usepackage{float}
\usepackage{arydshln}
\mathchardef\mhyphen="2D
\usepackage{multirow}
\usepackage[section]{placeins}

\copyrightyear{2021}
\acmYear{2021}
\setcopyright{acmcopyright}
\acmConference[AIES '21] {Proceedings of the 2021 AAAI/ACM Conference on AI, Ethics, and Society}{May 19--21, 2021}{Virtual Event, USA.}
\acmBooktitle{Proceedings of the 2021 AAAI/ACM Conference on AI, Ethics, and Society (AIES '21), May 19--21, 2021, Virtual Event, USA}
\acmPrice{15.00}
\acmISBN{978-1-4503-8473-5/21/05}
\acmDOI{10.1145/3461702.3462557}

\begin{document}
\fancyhead{}
\title{Measuring Model Biases in the Absence of Ground Truth}

\author{Osman Aka}
\affiliation{%
 \institution{Google}
 \country{U.S.A.}
}
\email{osmanaka@google.com}
\authornote{Equal contribution.}

\author{Ken Burke}
\affiliation{%
 \institution{Google}
  \country{U.S.A.}
}
\email{kenburke@google.com}
\authornotemark[1]

\author{Alex Bäuerle}
\affiliation{%
 \institution{Ulm University}
  \country{Germany}
}
\authornote{Work conducted during internship at Google.}

\author{Christina Greer}
\affiliation{%
\institution{Google}
 \country{U.S.A.}
}
 
\author{Margaret Mitchell}
\authornote{Work conducted while author was at Google.}
\affiliation{
\institution{Ethical AI LLC}
\country{U.S.A.}}









 


\begin{abstract}
The measurement of bias in machine learning often focuses on model performance across identity subgroups (such as $man$ and $woman$) with respect to groundtruth labels \cite{hardt2016equality}. However, these methods do not directly measure the \textit{associations} that a model may have learned, for example \textit{between} labels and identity subgroups. Further, measuring a model's bias requires a fully annotated evaluation dataset which may not be easily available in practice.

We present an elegant mathematical solution that tackles both issues simultaneously, using image classification as a working example. By treating a classification model's predictions for a given image as a set of labels analogous to a “bag of words” \cite{Jurafsky2009}, we rank the biases that a model has learned with respect to different identity labels. We use $\{man, woman\}$ as a concrete example of an identity label set (although this set need not be binary), and present rankings for the labels that are most biased towards one identity or the other. We demonstrate how the statistical properties of different association metrics can lead to different rankings of the most ``gender biased" labels, and conclude that normalized pointwise mutual information ($nPMI$) is most useful in practice. Finally, we announce an open-sourced $nPMI$ visualization tool using TensorBoard. 
\end{abstract}



\maketitle
\section{Introduction}

The impact of algorithmic bias in computer vision models has been well-documented \citep[c.f.,][]{pmlr-v81-buolamwini18a,ACLURekognition}. Examples of the negative fairness impacts of machine learning models include decreased pedestrian detection accuracy
on darker skin tones \cite{DBLP:journals/corr/abs-1902-11097}, gender stereotyping in image captioning \cite{DBLP:journals/corr/abs-1803-09797}, and perceived racial identities impacting unrelated labels \cite{stock2018convnets}. Many of these examples are directly related to currently deployed technology, which highlights the urgency of solving these fairness problems as adoption of these technologies continues to grow.

Many common metrics for quantifying fairness in machine learning models, such as Statistical Parity \cite{DBLP:journals/corr/abs-1104-3913}, Equality of Opportunity \cite{hardt2016equality} and Predictive Parity \cite{chouldechova2016fair}, rely on datasets with a significant amount of ground truth annotations for each label under analysis. However, some of the most commonly used datasets in computer vision have relatively sparse ground truth \cite{OpenImages}. One reason for this is the significant growth in the number of predicted labels. The benchmark challenge dataset PASCAL VOC introduced in 2008 had only 20 categories \cite{PascalRetro}, while less than 10 years later, the benchmark challenge dataset ImageNet provided hundreds of categories \cite{ILSVRC15}. As systems have rapidly improved, it is now common to use the full set of ImageNet categories, which number more than 20,000 \cite{imagenet_cvpr09,ImageNet}. 

While large label spaces offer a more fine-grained ontology of the visual world, they also increase the cost of implementing groundtruth-dependent fairness metrics. This concern is compounded by the common practice of collecting training datasets from multiple online resources \cite{COCO}. This can lead to patterns where specific labels are omitted in a biased way, either through human bias (e.g., crowdsourcing where certain valid labels or tags are omitted systematically) or through algorithmic bias (e.g., selecting labels for human verification based on the predictions of another model \cite{OpenImages}). If the ground truth annotations in a sparsely labelled dataset are potentially biased, then the premise of a fairness metric that ``normalizes" model prediction patterns to groundtruth patterns may be incomplete. In light of this difficulty, we argue that it is important to develop bias metrics that do not explicitly rely on ``unbiased'' ground truth labels.

In this work, we introduce a novel approach for measuring problematic model biases, focusing on the associations between model predictions directly.  This has several advantages compared to common fairness approaches in the context of large label spaces, making it possible to identify biases after the regular practice of running model inference over a dataset. We study several different metrics that measure associations between labels, building upon work in Natural Language Processing \cite{church-hanks-1990-word} and information theory. We perform experiments on these association metrics using the Open Images Dataset \cite{OpenImages} which has a large enough label space to illustrate how this framework can be generally applied, but we note that the focus of this paper is on introducing the relevant techniques and do not require any specific dataset.  We demonstrate that normalized pointwise mutual information ($nPMI$) is particularly useful for detecting the associations between model predictions and sensitive identity labels in this setting, and can uncover stereotype-aligned associations that the model has learned.  This metric is particularly promising because:
\begin{itemize}
    \item It requires no ground truth annotations.
    \item It provides a method for uncovering biases that the model itself has learned.
    \item It can be used to provide insight into per-label associations between model predictions and identity attributes.
    \item Biases for both low- and high-frequency labels are able to be detected and compared.
\end{itemize}
 Finally we announce an open-sourced visualization tool in TensorBoard that allows users to explore patterns of label bias in large datasets using the $nPMI$ metric.

\section{Related Work}

In 1990, Church and Hanks \cite{church-hanks-1990-word} introduced a novel approach to quantifying associations between words based on mutual information \cite{journals/bstj/Shannon48,fano1961transmission} and inspired by psycholinguistic work on word norms \cite{WordNorms} that catalogue words that people closely associate. For example, subjects respond more quickly to the word {\it nurse} if it follows a highly associated word such as {\it doctor} \cite{IAT,Caliskan183}. Church and Hanks' proposed metric applies mutual information to words using a {\it pointwise approach}, measuring co-occurrences of distinct word pairs rather than averaging over all words. This enables a quantification of the question, "How closely related are these words?" by measuring their co-occurrence rates relative to chance in the dataset of interest. In this case, the dataset of interest is a computer vision evaluation dataset, and the words are the labels that the model predicts.

This information theoretic approach to uncovering word associations became a prominent method in the field of Natural Language Processing, with applications ranging from measuring topic coherence \cite{aletras-stevenson-2013-evaluating} to collocation extraction \cite{Bouma2009NormalizedM} to great effect, although often requiring a good deal of preprocessing in order to incorporate details of a sentence's syntactic structure. However, without preprocessing, this method functions to simply measure word associations regardless of their order in sentences or relationship to one another, treating words as an unordered set of tokens (a so-called "bag-of-words") \cite{harris54}. 

As we show, this simple approach can be newly applied to an emergent problem in the machine learning ethics space: The identification of problematic associations that an ML model has learned. This approach is comparable to measuring correlations, although the common correlation metric of Spearman Rank  \cite{spearman04} operates on assumptions that are not suitable for this task, such as linearity and monotonicity.  The related correlation metric of 
 the Kendall Rank Correlation \cite{kendall} does not require such behavior, and we include comparisons with this approach. 
 
 Additionally, many potentially applicable metrics for this problem rely on simple counts of paired words, which does not take into consideration how the words are distributed with other words (e.g., sentence syntax or context); we will elaborate on how this information can be formally incorporated into a bias metric in the Discussion and Future Work sections.

This work is motivated by recent research on fairness in machine learning (e.g., \cite{hardt2016equality}), which at a high level seeks to define criteria that result in equal outcomes across different subpopulations. The focus in this paper is complementary to previous fairness work, honing in on ways to identify and quantify the specific problematic associations that a model may learn rather than providing an overall measurement of a model's unfairness. It also offers an alternative to fairness metrics that rely on comprehensive ground truth labelling, which is not always available for large datasets. 

\paragraph{Open Images Dataset.} The Open Images dataset we use in this work was chosen because it is open-sourced, with millions of diverse images and a large label space of thousands of visual concepts (see \cite{OpenImages} for more details). Furthermore, the dataset comes with pre-computed labels generated by a non-trivial algorithm that combines machine-generated predictions and human-verification; this allowed us to focus on analysis of label associations (rather than training a new classifier ourselves) and uncover the most common concepts related to sensitive characteristics, which in this dataset are ${man}$ and ${woman}$.

We now turn to a formal description of the problem we seek to solve.

\section{Problem Definition}

\begin{figure*}[h!]
\begin{adjustbox}{width=\linewidth,center}
\begin{tabular}{cc}
    \includegraphics[width=5.8cm]{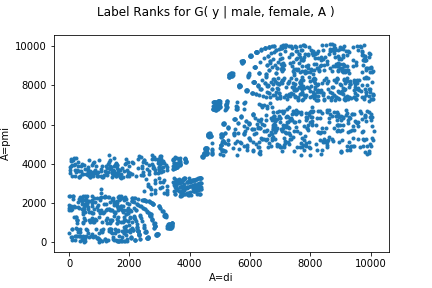} & 
    \includegraphics[width=5.8cm]{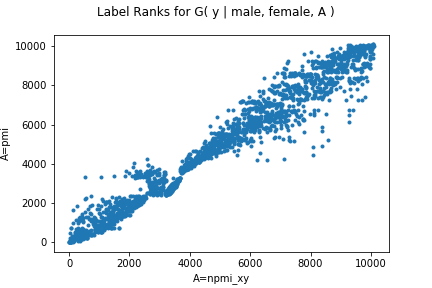} \\
    \multicolumn{2}{c}{\includegraphics[width=5.8cm]{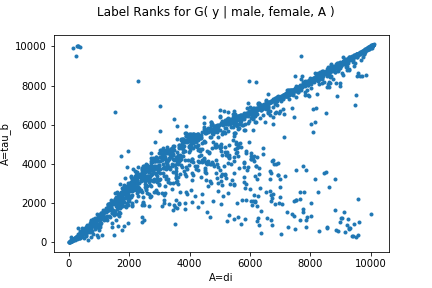}} \\
\end{tabular}
\end{adjustbox}
    \caption{Label ranking shifts for metric-to-metric comparison.     Each point represents the rank of a single label when sorted by $G(y | x_1, x_2, A(\cdot))$ (i.e., the label rank by gap). The coordinates represent the rankings by gap for different association metrics $A(\cdot)$ on the $x$ and $y$ axes. A highly-correlated plot along $y=x$ would imply that the two metrics lead to very similar bias rankings according to $G(\cdot)$.}
    \label{fig:label_rank_shift}

\end{figure*}

We have a dataset $\mathcal{D}$ which contains image examples and labels generated by a image classifier. This classifier takes one image example and predicts ``Is label $y_i$ relevant to the image?" for each label in $\mathcal{L}=\{y_1, y_2, ..., y_n\}$\footnote{For ease of notation, we use $y$ and $x$ rather than $\hat{y}$ and $\hat{x}$.}. We infer $P(y_i)$ and $P(y_i, y_j)$ from $\mathcal{D}$ such that for a given random image in $\mathcal{D}$, $P(y_i)$ is the probability of having $y_i$ as positive prediction and $P(y_i, y_j)$ is the joint probability of having both $y_i$ and $y_j$ as positive predictions. We further assume that we have identity labels  $x_1, x_2, ... x_n \in \mathcal{L}$ that belong to some sensitive identity group for which we wish to compute a bias metric (e.g., $man$ and $woman$ as labels for gender)\footnote{For the rest of the paper, we focus on only two identity labels with notation $x_1$ and $x_2$ for simplicity, however the identity labels need not be binary (or one-dimensional) in this way. The remainder of this work is straightforwardly extended to any number of identity labels by using the pairwise comparison of all identity labels, or by using a one-vs-all gap (e.g., $A(x, y) -\mathop{\mathbb{E}}{[A(x', y)]}$ where $x'$ is the set of all other $x$).}. These identity labels may even be generated by the same process that generates the labels $y$, as in our case where $man$, $woman$ and all other $y$ are elements of $\mathcal{L}$ predicted by the classifier. We then measure bias with respect to these identity labels for all other labels $y \in \mathcal{L}$. As the size of this label space $|\mathcal{L}|$ approaches tens of thousands and increases year by year for modern machine learning models, it is important to have a simple bias metric that can be computed and reasoned about at scale.

For ease of discussion in the rest of this paper, we denote any generic association metric $A(x_j, y)$, where $x_j$ is an identity label and $y$ is any other label. We define an {\it association gap} for label $y$ between two identity labels $[x_1, x_2]$ with respect to the association metric $A(x_j, y)$ as $G(y | x_1, x_2, A(\cdot)) = A(x_1, y) - A(x_2, y)$. For example, the association between the labels $woman$ and $bike$, $A(woman, bike)$ can be compared to the association between the labels $man$ and $bike$, $A(man, bike)$.  The difference between them is the \textbf{association gap} for the label $bike$: \\\\ $G(bike|woman,man,A(\cdot))$ =\\ \indent $A(woman, bike)$ - $A(man, bike)$ \\\\
We use this association gap $G(\cdot)$ as a measurement of ``bias" or ``skew" of a given label across sensitive identity subgroups.

The first objective we are interested in is the question ``Is the prediction of label $y$ biased towards either $x_1$ or $x_2$?". The second objective is then ranking the labels $y$ by the size of this bias. 
If $x_1$ and $x_2$ both belong to the same identity group (e.g., $man$ and $woman$ are both labels for ``gender"), then one may consider these measurements to approximate the \textit{gender bias}. 

We choose this gender example because of the abundance of these specific labels in the Open Images Dataset, however this choice should not be interpreted to mean that gender representation is one-dimensional, nor that paired labels are required for the general$^1$ approach. Nonetheless, this simplification is important because it allows us to demonstrate how a single per-label approximation of ``bias" can be measured between paired labels, and we leave details of further expansions, including calculations across multiple sensitive identity labels, to the Discussion section.

\subsection{Association Metrics}

We consider several sets of related association metrics $A(\cdot)$ that can be applied given the constraints of the problem at hand -- limited groundtruth, non-linearity, and limited assumptions about the underlying distribution of the data. All of these metrics share in common the general intuition of measuring how labels associate with each other in a dataset, but as we will demonstrate, they yield very different results due to differences in how they quantify this notion of ``association".

We first consider fairness metrics as types of association metrics.  One of the most common fairness metrics, Demographic (or Statistical) Parity \cite{DBLP:journals/corr/abs-1104-3913,berk2016primer,hardt2016equality}, a quantification of the legal doctrine of  Disparate Impact \cite{barocas2014datas}, can be applied directly for the given task constraints.\footnote{Other common fairness metrics, such as Equality of Opportunity, require both a model prediction and a groundtruth label, which makes the correct way to apply them to this task less clear. We leave this for further work.} Other metrics that are possible to adopt for this task include those based on Intersection-over-Union (IOU) measurements, and metrics based on correlation and statistical tests. We next describe these metrics in further detail and their relationship to the task at hand. In summary, we compare the following families of metrics:
\begin{itemize}
    \item Fairnesss: Demographic Parity ($DP$)
    \item Entropy: Pointwise Mutual Information ($PMI$), Normalized Pointwise Mutual Information ($nPMI$).
    \item IOU: Sørensen-Dice Coefficient ($SDC$), Jaccard Index ($JI$).
    \item Correlation and Statistical Tests: Kendall Rank Correlation ($\tau_b$), Log-Likelihood Ratio ($LLR$), $t$-test. 
\end{itemize}

 One of the important aspects of our problem setting is the counts of images with labels and label intersections, i.e., $C(y)$, $C(x_1, y)$, and $C(x_2,y)$. These values can span a large range for different labels $y$ in the label set $\mathcal{L}$, depending on how common they are in the dataset. Some metrics are theoretically more sensitive to the frequencies/counts of the label $y$ as determined by their nonzero partial derivatives with respect to $P(y)$ (see Table 2). However, as we further discuss in the Experiments and Discussion sections, our experiments indicate that in practice, metrics with non-zero partial derivatives are surprisingly better able to capture biases across a range of label frequencies than metrics with a zero partial derivative. Differential sensitivity to label frequency could be problematic in practice for two reasons:
\begin{enumerate}
    \item It would not be possible to compare $G(y | x_1, x_2, A(\cdot))$ {\it between} different labels $y$ with different marginal frequencies (counts) $C(y)$. For example, the ideal bias metric should be able to capture gender bias equally well for both $car$ and $Nissan$ $Maxima$ even though the first label is more common than the second.
    \item The alternative, bucketizing labels by marginal frequency and setting distinct thresholds per bucket, would add significantly more hyperparameters and essentially amount to manual frequency-normalization. 
\end{enumerate}

The following sections contain basic explanations of these metrics for a general audience, with the running example of $bike, man,$ and $woman$. We leave further mathematical analyses of the metrics to the Appendix. However, integral to the application of $nPMI$ in this task is the choice of normalization factor, and so we discuss this in further detail in the Normalizing $PMI$ subsection.
~\\~\\
\noindent
\textit{Demographic Parity}
\[G(y|x_1,x_2,DP)=P(y|x_1) - P(y|x_2)\]

\noindent Demographic Parity focuses on differences between the conditional probability of $y$ given $x_1$ and $x_2$: How likely $bike$ is for $man$ vs $woman$.
~\\

\noindent\textit{Entropy} \\
\[G(y|x_1,x_2,PMI)=ln\left(\frac{P(x_1,y)}{P(x_1)P(y)}\right) -
ln\left(\frac{P(x_2,y)}{P(x_2)P(y)}\right)\]
~\\
\noindent Pointwise Mutual Information, adapted from information theory, is the main entropy-based metric studied here. In this form, we are analyzing the entropy difference between $\left[x_1, y\right]$ and $\left[x_2, y\right]$. This essentially examines the dependence of, for example, the ${bike}$ label distribution on two other label distributions: ${man}$ and ${woman}$. 

~\\
\noindent
\textit{Remaining Metrics}

We use the Sørensen-Dice Coefficient ($SDC$),  which has the commonly-used F1-score as one of its variants; the Jaccard Index ($JI$), a common metric in Computer Vision also known as Intersection Over Union (IOU); Log-Likelihood Ratio ($LLR$), a classic flexible comparison approach; Kendall Rank Correlation, which is also known as $\tau_b$-correlation, and is the particular \textit{correlation} method that can be reasonably applied in this setting; and the $t\mhyphen test$, a common statistical significance test that can be adapted in this setting \cite{Jurafsky2009}. Each of these metrics have different behaviours, however, we limit our mathematical explanation to the Appendix, as we found these metrics are either less useful in practice or behave similarly to other metrics in this use case.

\subsection{Normalizing PMI}\label{sec:normalizing}
One major challenge in our problem setting is the sensitivity of these association metrics to the frequencies/counts of the labels in $\mathcal{L}$. Some metrics are weighted more heavily towards common labels (i.e., large marginal counts, $C(y)$) in spite of differences in their joint probabilities with identity labels ($P(x_1, y) , P(x_2, y)$). The opposite is true for other metrics, which are weighted towards rare labels with smaller marginal frequencies. In order to compensate for this problem, several different normalization techniques have been applied to $PMI$ \cite{Bouma2009NormalizedM,kdir11}. Common normalizations include:
\begin{itemize}
    \item $nPMI_y$: Normalizing each term by $P(y)$.
    \item $nPMI_{xy}$: Normalizing the two terms by $P(x_1,y)$ and $P(x_2, y)$, respectively.
    \item $PMI^2$: Using $P(x_1,y)^2$ and $P(x_2, y)^2$ instead of $P(x_1, y)$ and $P(x_2, y)$, the normalization effects of which are further illustrated in the Appendix.
\end{itemize}
Each of these normalization methods have different impacts on the $PMI$ metric. The main advantage of these normalizations is the ability to compare association gaps {\it between} label pairs $[y_1, y_2]$ (e.g., comparing the gender skews of two labels like \textit{Long Hair} and \textit{Dido Flip}) even if $P(y_1)$ and $P(y_2)$ are very different. In the Experiments section, we discuss which of these is most effective and meaningful for the fairness and bias use case motivating this work.

\section{Experiments}
In order to compare these metrics, we use the Open Images Dataset (OID) \cite{OpenImages} described above in the Related Works section. This dataset is useful for demonstrating realistic bias detection use cases because of the number of distinct labels that may be applied to the images (the dataset itself is annotated with nearly 20,000 labels). The label space is also diverse, including objects ({\it cat}, {\it car}, {\it tree}), materials ({\it leather}, {\it basketball}), moments/actions ({\it blowing out candles}, {\it woman playing guitar}), and scene descriptors and attributes ({\it beautiful}, {\it smiling}, {\it forest}). We seek to measure the gender bias as described above for each of these labels, and compare these bias values directly in order to determine which of the labels are most skewed towards associating with the labels $man$ and $woman$.

In our experiments, we apply each of the association metrics $A(x_j, y)$ to the machine-generated predictions for identity labels $x_1=man, x_2=woman$ and all other labels $y$ in the Open Images Dataset. We then compute the gap value $G(\cdot)$ between the identity labels for each label $y$ and sort, providing a ranking of labels that are most biased towards $x_1$ or $x_2$. 
As we will show, sorting labels by this \textit{association gap} creates different rankings for different association metrics. We examine which labels are ranked within the top 100 for the different association metrics in the Top 100 Labels by Metric Gaps subsection.

\subsection{Label Ranks}\label{sec:ranks}

The first experiment we performed is to compute the association metrics and the gaps between them for different labels -- $A(x_1, y)$, $A(x_2, y)$ and $G(y | x_1, x_2, A(\cdot))$ -- over the OID dataset. We then sorted the labels by $G(\cdot)$ and studied how the ranking by gap differed between different association metrics $A(\cdot)$. Figure \ref{fig:label_rank_shift} shows examples of these metric-to-metric comparisons of label rankings by gap (all other metric comparisons can be found in the Appendix). We can see that a single label can have quite a different ranking depending on the metric.

When comparing metrics, we found that they grouped together in a few clusters based on similar ranking patterns when sorting by $G(y | man, woman, A(\cdot))$. In the first cluster, pairwise comparisons between $PMI$, $PMI^2$ and $LLR$ show linear relationships when sorting labels by $G(\cdot)$. Indeed, while some labels show modest changes in rank between these metrics, they share {\it all} of their top 100 labels, and $>99\%$ of label pairs maintain the same relative ranking between metrics. By contrast, there are only $7$ labels in common between the top 100 labels of $PMI^2$ and $SDC$. Due to the similar behavior of this cluster of metrics, we chose to focus on $PMI$ as representative of these 3 metrics moving forward (see the Appendix for further details on these relationships).

\begin{table}
\footnotesize
\resizebox{\linewidth}{!}{%
\begin{tabular}{ |@{\hspace{.2em}}c@{\hspace{.1em}}|@{\hspace{.2em}}c@{\hspace{.2em}}|@{\hspace{.2em}}c@{\hspace{.2em}}|@{\hspace{.25em}}c@{\hspace{.25em}}|  }
\hline
{Metrics}&Min/Max&Min/Max&Min/Max\\
& $C(y)$ & $C(x_1,y)$ &  $C(x_2, y)$\\
\hline
&&&\\
$PMI$&       15 / 10,551    &        1 / 1,059     &        8 / 7,755     \\[2ex]
$PMI^2$&       15 / 10,551    &        1 / 1,059     &        8 / 7,755     \\[2ex]
$LLR$&       15 / 10,551    &        1 / 1,059     &        8 / 7,755     \\[2ex]
$DP$&     ~6,104 / 785,045   &      ~628 / 239,950   &     5,347 / 197,795   \\[2ex]
$JI$&     ~4,158 / 562,445   &      ~399 / 144,185   &     3,359 / 183,132   \\[2ex]
$SDC$&    ~2,906 / 562,445   &      139 / 144,185   &     2,563 / 183,132   \\[2ex]
$nPMI_{y}$&       35 / 562,445   &        1/144,185   &        9 / 183,132   \\[2ex]
$nPMI_{xy}$&       34 / 270,748   &        1 / 144,185   &       20 / 183,132   \\[2ex]
$\tau_b$&     ~6,104 / 785,045   &      ~628 / 207,723   &     ~5,347 / 183,132   \\[2ex]
$t \mhyphen test$&      960 / 562,445   &       72 / 144,185   &      870 / 183,132   \\[1ex]
\hline
\end{tabular}%
}
\caption{Minimum and maximum counts $C(y)$ of the top 100 labels with the largest association gaps for each metric. Note that these min/max values for $C(y)$ vary by orders of magnitude for different metrics. A larger version of this table is available in Appendix, Table \ref{app_minmaxcounts}.}
\label{fig:table_of_counts}
\end{table}

\begin{figure*}[htp]
    \centering
\resizebox{\linewidth}{!}{%
    \begin{tabular}{ccc}
         \includegraphics{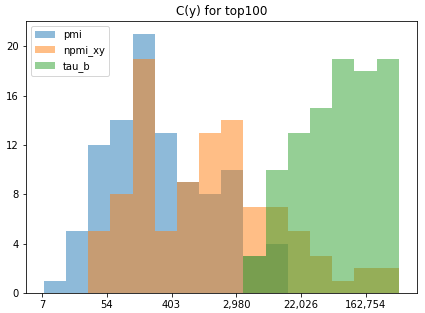} &
        \includegraphics{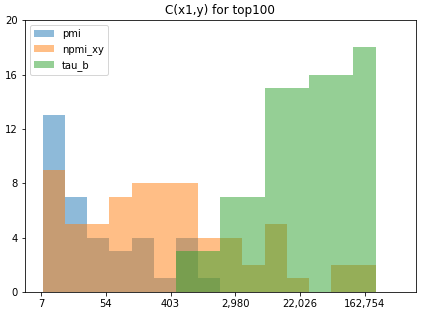} &
        \includegraphics{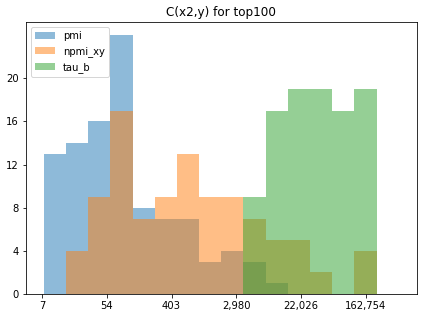}   
    \end{tabular}%
    }
    
    \caption{Top 100 count distributions for $PMI$, $nPMI_{xy}$, and $\tau_b$}
    \label{fig:top_100_count_dist}
    The distribution of $C(y)$, $C(x_1,y)$, and $C(x_2,y)$ for the top 100 labels sorted by gap $G(y|x_1,x_2,A(\cdot))$ for $PMI$, $nPMI_{xy}$, and $\tau_b$. The $x$-axis is the logarithmic-scaled bins and the $y$-axis is the number of labels which have the corresponding count values in that bin.
\end{figure*}

Similar results were obtained for another cluster of metrics: $DP$, $JI$, and $SDC$. All pairwise comparisons generated linear plots, with about 70\% of the top 100 labels shared in common between these metrics when sorted by $G(\cdot)$. Furthermore, about 95\% of pairs of those overlapping labels maintained the same relative ranking between metrics. Similar to the $PMI$ cluster, we chose to focus on Demographic Parity ($DP$) as the representative metric from its cluster, due to its mathematical simplicity and prominence in fairness literature.

We next sought to understand how incremental changes to the counts of the labels, $C(y)$, affect these association gap rankings in a real dataset (see Appendix Section \ref{plot:movement}). To achieve this, we added a fake label to the real labels of OID, setting  initial values for its counts and co-occurences in the dataset, $P(y)$, $P(x_1, y)$, and $P(x_2, y)$. Then we incrementally increased or decreased the count of label $y$, and measured whether its bias ranking in $G$ would change relative to the other labels in OID. We repeated this procedure for different orders of magnitude of label count $C(y)$ while maintaining the ratio $P(x_1,y)/P(x_2,y)$ as constant. 
\begin{table*}
\begin{adjustbox}{width=2.2\columnwidth,center}
\begin{tabular}{ |c|c|c|c|  }
\hline
&$\partial p(y)$&$\partial p(x_1, y)$&$\partial p(x_2, y)$\\
\hline
&&&\\
$\partial$DP&$0$&$\frac{1}{p(x_1)}$&$\frac{-1}{p(x_2)}$ \\ [3ex] 

 $\partial$PMI&$0$&$\frac{1}{p(x_1,y)}$&$\frac{-1}{p(x_2,y)}$ \\[3ex] 
$\partial nPMI_{y}$&$\frac{ln(\frac{p(x_2|y)}{p(x_1|y)})}{ln^2(p(y))p(y)}$&$\frac{1}{ln(p(y))p(x_1, y)}$&$\frac{-1}{ln(p(y))p(x_2, y)}$\\[3ex] 
$\partial nPMI_{xy}$&$\frac{1}{ln(p(x_1,y))p(y)}-\frac{1}{ln(p(x_2,y))p(y)}$&$\frac{ln(p(y))-ln(p(x_1))}{ln^2(p(x_1,y))p(x_1,y)}$&$\frac{ln(p(x_2))-ln(p(y))}{ln^2(p(x_2,y))p(x_2,y)}$\\[3ex]
$\partial PMI^2$&0&$\frac{2}{p(x_1,y)}$&$\frac{-2}{p(x_2,y)}$\\[3ex] 
$\partial$SDC&\text{\small\textit{{see Appendix \ref{app:orientation_details}}}}&$\frac{1}{p(x_1)+p(y)}$&$\frac{-1}{p(x_2)+p(y)}$\\[3ex]
$\partial$JI&\text{\small{\textit{see Appendix \ref{app:orientation_details}}}}&$\frac{p(x_1)+p(y)}{(p(x_1)+p(y)-p(x_1,y))^2}$&$\frac{p(x_2)+p(y)}{(p(x_2)+p(y)-p(x_2,y))^2}$\\[3ex]
$\partial$LLR&$0$&$\frac{1}{p(x_1, y)}$&$\frac{-1}{p(x_2, y)}$\\[3ex]
$\partial \tau_b$&\text{\small{\textit{see Appendix \ref{app:orientation_details}}}}&$\frac{(2-\frac{4}{n})}{\sqrt{(p(x_1)-p(x_1)^2)(p(y)-p(y)^2)}}$&$\frac{(\frac{4}{n}-2)}{\sqrt{(p(x_2)-p(x_2)^2)(p(y)-p(y)^2)}}$   \\[3ex]
$\partial t \mhyphen  test\_gap$&$\frac{\sqrt{p(x_2)}-\sqrt{p(x_1)}}{2\sqrt{p(y)}}$&$\frac{1}{\sqrt{p(x_1)p(y)}}$&$\frac{-1}{\sqrt{p(x_2)p(y)}}$ \\ [3ex]
\hline
\end{tabular}
\end{adjustbox}
\caption{Metric orientations.}
\label{fig:table_of_orientations}
This table shows the partial derivatives of the metrics with respect to $P(y)$, $P(x_1,y)$, and $P(x_2,y)$. This provides a quantification of the theoretical sensitivity of the metrics for different probability values of $P(y)$, $P(x_1,y)$, and $P(x_2,y)$. We see similar experimental results for the metrics with similar orientations (see Appendix).
\end{table*}

This experiment allowed us to determine whether the theoretical sensitivities of each metric to label frequency $P(y)$, as determined by partial derivatives $\partial{A(\cdot)}/\partial{P(y)}$ (see Table \ref{fig:table_of_orientations}), would hold in the context of real-world data, where the underlying distribution of label frequencies may not be uniform. If certain subregions of the label distribution are relatively sparse, for example, then the gap ranking of our hypothetical label may not change even if  $\partial{A(\cdot)}/\partial{C(y)}\neq0$. However, in practice we do not observe this behavior in the tested settings (see Appendix for plots of these experiments), where label rank moves with label count roughly as predicted by the partial derivatives in Table \ref{fig:table_of_orientations}. In fact, we observed that metrics with larger partial derivatives for $x_1$, $x_2$, or $y$ often led to a larger change in rank. For example, slightly increasing $P(x_1,y)$ when $y$ always co-occurs with $x_1$, $P(y)=P(x_1,y)$ affects ranking more for $A=nPMI_y$ compared to $A=PMI$ (see Appendix).

\subsection{Top 100 Labels by Metric Gaps}

When applying these metrics to fairness and bias use cases, model users may be most interested in surfacing the labels with the largest association gaps. If one filters results to a ``top K" label set, then the normalization chosen could lead to vastly different sets of labels (e.g., as mentioned earlier, $PMI^2$ and $SDC$ only shared 7 labels in their top 100 set for OID).

To further analyze this issue, we calculated simple values for each metric's top 100 labels sorted by $G(\cdot)$: minimum and maximum values of $C(y)$, $C(x_1,y)$ and $C(x_2,y)$ as shown in Table 1. The most salient point is that the clusters of metrics from the Label Ranks subsection also appear to hold in this analysis as well; $PMI$, $PMI^2$, and $LLR$ have low $C(y)$, $C(x_1,y)$ and $C(x_2,y)$ ranges, whereas $DP$, $JI$, and $SDC$ have relatively high ranges. Another straightforward observation we can make is that the $nPMI_y$ and $nPMI_{xy}$ ranges are much broader than the first two clusters, and include the other metrics' ranges especially for the joint co-occurrences, $C(x_1,y)$ and $C(x_2,y)$.

To demonstrate this point more clearly, we plot the distributions of these counts for $PMI$, $nPMI_{xy}$ and $\tau_b$ skews in Figure \ref{fig:top_100_count_dist} (all other combinations can be found in the Appendix). These three metric distributions show that gap calculations based on $PMI$ (blue distribution) exclusively rank labels with low counts in the top 100 most skewed labels, where $\tau_b$ calculations (green distribution) almost exclusively rank labels with much higher counts. The exception is $nPMI_y$ and $nPMI_{xy}$ (orange distribution); these two metrics are capable of capturing labels across a range of marginal frequencies. In other words, ranking labels by $PMI$ gaps is likely to highlight \textit{rare labels}, ranking by $\tau_b$ will highlight \textit{common labels}, and ranking by $nPMI_{xy}$ will highlight \textit{both rare and common labels}.

An example of this relationship between association metric choice and label commonality can be seen in Table \ref{tab:top_15_examples} (note, here we use Demographic Parity instead of $\tau_b$ because they behave similarly in this respect). In this table, we show the ``Top 15 labels" most heavily skewed towards $woman$ relative to $man$ according to $DP$, unnormalized $PMI$, and $nPMI_{xy}$. $DP$ almost exclusively highlights common labels predicted for over 100,000 images in the dataset (e.g., \textit{Happiness} and \textit{Fashion}), whereas $PMI$ largely highlights rarer labels predicted for less than 1,000 images (e.g., \textit{Treggings} and \textit{Boho-chic}). By contrast, $nPMI_{xy}$ highlights both common and rare labels (e.g., \textit{Long Hair} as well as \textit{Boho-chic}).

\section{Discussion}\label{sec:discussion}

\begin{table*}[htp]
\centering
\begin{tabular}{ |c||c|c|c|c|c|c|  }
\hline
\multicolumn{1}{|c}{Metric $A$} &\multicolumn{2}{c}{$DP$}&\multicolumn{2}{c}{$PMI$}&\multicolumn{2}{c|}{$nPMI_{xy}$} \\
\hline
Ranks&Label $y$ &Count&Label $y$ &Count&Label $y$ &Count\\
\hline
0&&265,853&Dido Flip&140&&610\\
1&&270,748&Webcam Model&184&Dido Flip&140\\
2&&221,017&Boho-chic&151&&2,906\\
3&&166,186&&610&Eye Liner&3,144\\
4&Beauty&562,445&Treggings&126&Long Hair&56,832\\
5&Long Hair&56,832&Mascara&539&Mascara&539\\
6&Happiness&117,562&&145&Lipstick&8,688\\
7&Hairstyle&145,151&Lace Wig&70&Step Cutting&6,104\\
8&Smile&144,694&Eyelash Extension&1,167&Model&10,551\\
9&Fashion&238,100&Bohemian Style&460&Eye Shadow&1,235\\
10&Fashion Designer&101,854&&78&Photo Shoot&8,775\\
11&Iris&120,411&Gravure Idole&200&Eyelash Extension&1,167\\
12&Skin&202,360&&165&Boho-chic&460\\
13&Textile&231,628&Eye Shadow&1,235&Webcam Model&151\\
14&Adolescence&221,940&&156&Bohemian Style&184\\
\hline
\end{tabular}
\caption{Top 15 labels skewed towards $woman$.}
\label{tab:top_15_examples}
Ranking by the gap of label $y$ between $woman$  $x_1$ and $man$ $x_2$, according to $G(y|x_1,x_2,A(\cdot))$. Identity label names are omitted.
\end{table*}

In the previous section, we first showed that some association metrics behave very similarly when ranking labels from the Open Images Dataset (OID). We then showed that the mathematical orientations and sensitivity of these metrics align with experimental results from OID. Finally, we showed that the different normalizations affect whether labels with high or low marginal frequencies are likely to be detected as having a significant bias according to $G(y | x_1, x_2, A(\cdot))$ in this dataset. We arrive at the conclusion that the $nPMI$ metrics are preferable to other commonly used association metrics in the problem setting of detecting biases without groundtruth labels.

What is the intuition behind this particular association metric as a bias metric? All of the studied entropy-based metrics (gaps in $nPMI$ as well as $PMI$) approximately correspond to whether one identity label $x_1$ co-occurs with the target label $y$ more often than another identity label $x_2$ \textbf{relative to chance levels}. This chance-level normalization is important because even completely unbiased labels would still co-occur at some baseline rate by chance alone.

The further normalization of $PMI$ by either the marginal or joint probability ($nPMI_{y}$ and $nPMI_{xy}$, respectively) takes this one step further in practice by surfacing labels with larger marginal counts at higher ranks in $G(y | x_1, x_2, nPMI)$ alongside labels with smaller marginal counts. This is a somewhat surprising result, because in theory $PMI$ should already be independent of the marginal frequency of $P(y)$ (because $\partial{PMI}/\partial{p(y)}=0$), whereas this derivative for $nPMI$ is non-zero. When we examined this pattern in practice, the labels with smaller counts can achieve very large $P(x_1,y)$/$P(x_2,y)$ ratios (and therefore their bias rankings can get very high) merely by reducing the denominator to a single image example. $PMI$ is unable to compensate for this noise, whereas the normalizations we use for $nPMI$ allow us to capture a significant amount of common labels in the top 100 labels by $G(y|x_1,x_2,nPMI)$ in spite of this pattern. This result is indicated by the ranges in Table \ref{fig:table_of_counts} and Figure \ref{fig:top_100_count_dist}, as well as the set of both common and rare labels for $nPMI$ in Table \ref{tab:top_15_examples}. 

Indeed, if the evaluation set is properly designed to match the distribution of use cases of a classification model ``in the wild", then we argue more common labels that have a smaller $P(x_1,y)$/$P(x_2,y)$ ratio are still critical to audit for biases. Normalization strategies must be titrated carefully to balance this simple ratio of joint probabilities with the label's rarity in the dataset.

An alternative solution to this problem could be bucketing labels by their marginal frequency. We argue this is a suboptimal solution for two reasons. First, determining even a single threshold hyperparameter is a painful process for defining fairness constraints. Systems that prevent models from being published if their fairness discrepancies exceed a threshold would then be required to titrate this threshold for every bucket. Secondly, bucketing labels by frequency is essentially a manual and discontinuous form of normalization; we argue that building normalization into the metric directly is a more elegant solution.

\begin{figure*}[htp]
    \centering

    \includegraphics[width=\linewidth]{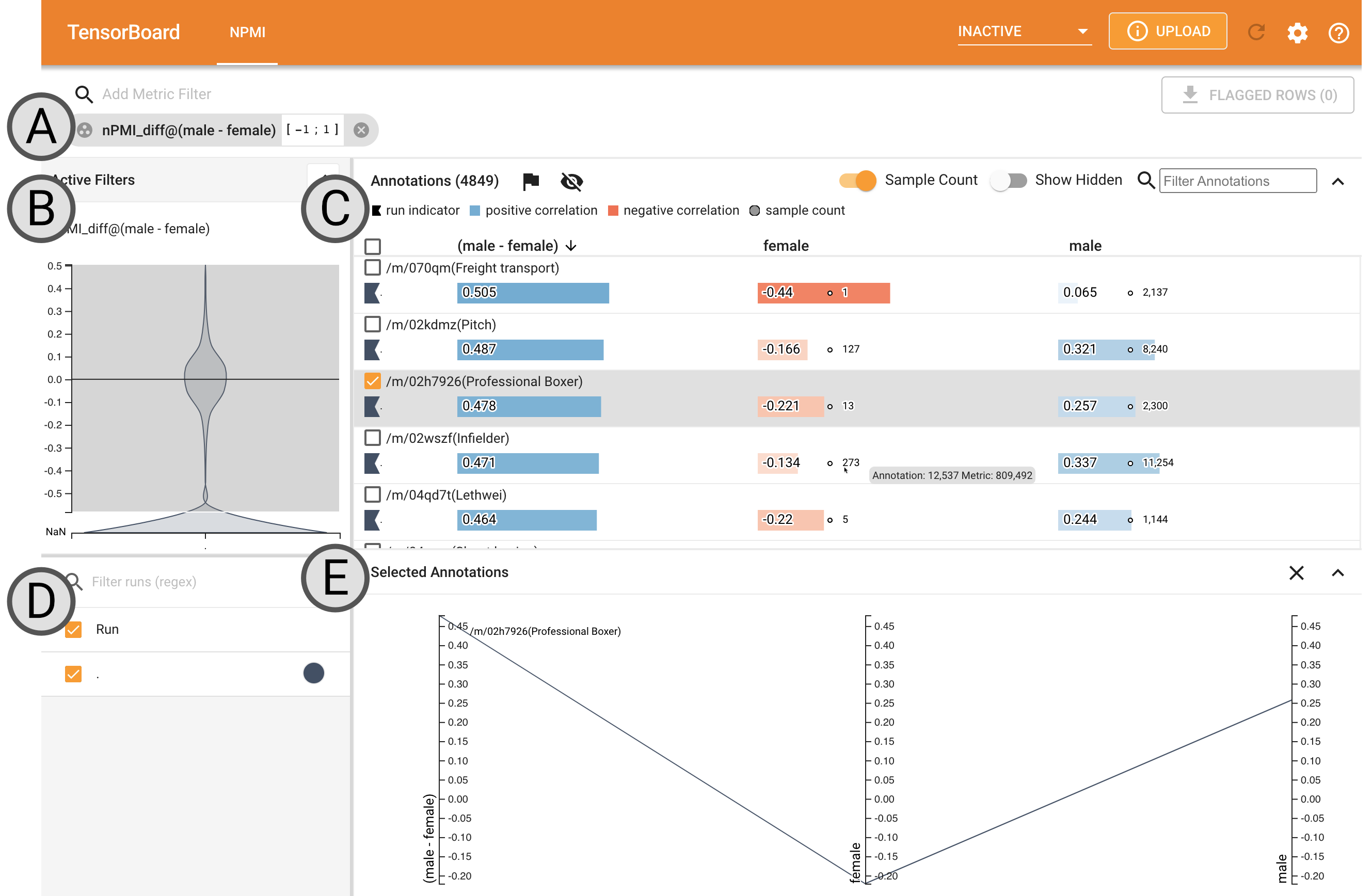}
    \caption{Open-sourced tool we implemented for bias investigation.}
    \label{fig:plugin}
    In \textbf{A}, annotations are filtered by an association metric, in this case the $nPMI$ difference ($nPMI\_{gap}$) between $male$ and $female$.
    In \textbf{B}, distribution of association values.
    In \textbf{C}, the filtered annotations view.
    In \textbf{D}, different models or datasets can be selected for display.
    In \textbf{E}, a parallel coordinates visualization of selected annotations to assess where association differences come from.
\end{figure*}

Finally, to enable detailed investigation of the model predictions, we implemented and open-sourced a tool to visualize $nPMI$ metrics as a TensorBoard plugin for developers\footnote{\url{https://github.com/tensorflow/tensorboard/tree/master/tensorboard/plugins/npmi}} (see Figure \ref{fig:plugin}). It allows users to investigate discrepancies between two or more identity labels and their pairwise comparisons. Users can visualize probabilities, image counts, sample and label distributions, and filter, flag, and download these results.

\section{Conclusion and Future Work}

In this paper we have described association metrics that can measure \textit{gaps} -- biases or skews -- towards specific labels in the large label space of current computer vision classification models. These metrics do not require ground truth annotations, which allows them to be applied in contexts where it is difficult to apply standard fairness metrics such as Equality of Opportunity \cite{hardt2016equality}. According to our experiments, Normalized Pointwise Mutual Information ($nPMI$) is a particularly useful metric for measuring specific biases in a real-world dataset with a large label space, e.g., the Open Images Dataset.

This paper also introduces several questions for future work. The first is whether $nPMI$ is also similarly useful as a bias metric in small label spaces (e.g., credit and loan applications). Second, if we were to have exhaustive ground truth labels for such a dataset, how would the sensitivity of $nPMI$ in detecting biases compare to ground-truth-dependent fairness metrics? Finally, in this work we treated the labels predicted for an image as a flat set. However, just like sentences have rich syntactic structure beyond the ``bag-of-words" model in NLP, images also have rich structure and relationships between objects that are not captured by mere rates of binary co-occurrence. This opens up the possibility that {\it within-image label relationships} could be leveraged to better understand how concepts are associated in a large computer vision dataset. We leave these questions for future work.

\bibliographystyle{ACM-Reference-Format}
\balance
\bibliography{bibliography}


\begin{thebibliography}{29}


\ifx \showCODEN    \undefined \def \showCODEN     #1{\unskip}     \fi
\ifx \showDOI      \undefined \def \showDOI       #1{#1}\fi
\ifx \showISBNx    \undefined \def \showISBNx     #1{\unskip}     \fi
\ifx \showISBNxiii \undefined \def \showISBNxiii  #1{\unskip}     \fi
\ifx \showISSN     \undefined \def \showISSN      #1{\unskip}     \fi
\ifx \showLCCN     \undefined \def \showLCCN      #1{\unskip}     \fi
\ifx \shownote     \undefined \def \shownote      #1{#1}          \fi
\ifx \showarticletitle \undefined \def \showarticletitle #1{#1}   \fi
\ifx \showURL      \undefined \def \showURL       {\relax}        \fi
\providecommand\bibfield[2]{#2}
\providecommand\bibinfo[2]{#2}
\providecommand\natexlab[1]{#1}
\providecommand\showeprint[2][]{arXiv:#2}

\bibitem[\protect\citeauthoryear{Aletras and Stevenson}{Aletras and
  Stevenson}{2013}]%
        {aletras-stevenson-2013-evaluating}
\bibfield{author}{\bibinfo{person}{Nikolaos Aletras} {and}
  \bibinfo{person}{Mark Stevenson}.} \bibinfo{year}{2013}\natexlab{}.
\newblock \showarticletitle{Evaluating Topic Coherence Using Distributional
  Semantics}. In \bibinfo{booktitle}{\emph{Proceedings of the 10th
  International Conference on Computational Semantics ({IWCS} 2013) {--} Long
  Papers}}. \bibinfo{publisher}{Association for Computational Linguistics},
  \bibinfo{address}{Potsdam, Germany}, \bibinfo{pages}{13--22}.
\newblock
\urldef\tempurl%
\url{https://www.aclweb.org/anthology/W13-0102}
\showURL{%
\tempurl}


\bibitem[\protect\citeauthoryear{Barocas and Selbst}{Barocas and
  Selbst}{2014}]%
        {barocas2014datas}
\bibfield{author}{\bibinfo{person}{Solon Barocas} {and}
  \bibinfo{person}{Andrew~D. Selbst}.} \bibinfo{year}{2014}\natexlab{}.
\newblock \showarticletitle{{Big Data's Disparate Impact}}.
\newblock \bibinfo{journal}{\emph{SSRN eLibrary}} (\bibinfo{year}{2014}).
\newblock


\bibitem[\protect\citeauthoryear{Berk}{Berk}{2016}]%
        {berk2016primer}
\bibfield{author}{\bibinfo{person}{Richard Berk}.}
  \bibinfo{year}{2016}\natexlab{}.
\newblock \showarticletitle{A primer on fairness in criminal justice risk
  assessments}.
\newblock \bibinfo{journal}{\emph{The Criminologist}} \bibinfo{volume}{41},
  \bibinfo{number}{6} (\bibinfo{year}{2016}), \bibinfo{pages}{6--9}.
\newblock


\bibitem[\protect\citeauthoryear{Bouma}{Bouma}{2009}]%
        {Bouma2009NormalizedM}
\bibfield{author}{\bibinfo{person}{G. Bouma}.} \bibinfo{year}{2009}\natexlab{}.
\newblock \showarticletitle{Normalized (pointwise) mutual information in
  collocation extraction}.
\newblock


\bibitem[\protect\citeauthoryear{Buolamwini and Gebru}{Buolamwini and
  Gebru}{2018}]%
        {pmlr-v81-buolamwini18a}
\bibfield{author}{\bibinfo{person}{Joy Buolamwini} {and}
  \bibinfo{person}{Timnit Gebru}.} \bibinfo{year}{2018}\natexlab{}.
\newblock \showarticletitle{Gender Shades: Intersectional Accuracy Disparities
  in Commercial Gender Classification} \emph{(\bibinfo{series}{Proceedings of
  Machine Learning Research}, Vol.~\bibinfo{volume}{81})},
  \bibfield{editor}{\bibinfo{person}{Sorelle~A. Friedler} {and}
  \bibinfo{person}{Christo Wilson}} (Eds.). \bibinfo{publisher}{PMLR},
  \bibinfo{address}{New York, NY, USA}, \bibinfo{pages}{77--91}.
\newblock
\urldef\tempurl%
\url{http://proceedings.mlr.press/v81/buolamwini18a.html}
\showURL{%
\tempurl}


\bibitem[\protect\citeauthoryear{Burns, Hendricks, Darrell, and Rohrbach}{Burns
  et~al\mbox{.}}{2018}]%
        {DBLP:journals/corr/abs-1803-09797}
\bibfield{author}{\bibinfo{person}{Kaylee Burns}, \bibinfo{person}{Lisa~Anne
  Hendricks}, \bibinfo{person}{Trevor Darrell}, {and} \bibinfo{person}{Anna
  Rohrbach}.} \bibinfo{year}{2018}\natexlab{}.
\newblock \showarticletitle{Women also Snowboard: Overcoming Bias in Captioning
  Models}.
\newblock \bibinfo{journal}{\emph{CoRR}}  \bibinfo{volume}{abs/1803.09797}
  (\bibinfo{year}{2018}).
\newblock
\showeprint[arxiv]{1803.09797}
\urldef\tempurl%
\url{http://arxiv.org/abs/1803.09797}
\showURL{%
\tempurl}


\bibitem[\protect\citeauthoryear{Caliskan, Bryson, and Narayanan}{Caliskan
  et~al\mbox{.}}{2017}]%
        {Caliskan183}
\bibfield{author}{\bibinfo{person}{Aylin Caliskan}, \bibinfo{person}{Joanna~J.
  Bryson}, {and} \bibinfo{person}{Arvind Narayanan}.}
  \bibinfo{year}{2017}\natexlab{}.
\newblock \showarticletitle{Semantics derived automatically from language
  corpora contain human-like biases}.
\newblock \bibinfo{journal}{\emph{Science}} \bibinfo{volume}{356},
  \bibinfo{number}{6334} (\bibinfo{year}{2017}), \bibinfo{pages}{183--186}.
\newblock
\showISSN{0036-8075}
\urldef\tempurl%
\url{https://doi.org/10.1126/science.aal4230}
\showDOI{\tempurl}
\showeprint{https://science.sciencemag.org/content/356/6334/183.full.pdf}


\bibitem[\protect\citeauthoryear{Chouldechova}{Chouldechova}{2016}]%
        {chouldechova2016fair}
\bibfield{author}{\bibinfo{person}{Alexandra Chouldechova}.}
  \bibinfo{year}{2016}\natexlab{}.
\newblock \bibinfo{title}{Fair prediction with disparate impact: A study of
  bias in recidivism prediction instruments}.
\newblock
\newblock
\showeprint[arxiv]{1610.07524}~[stat.AP]


\bibitem[\protect\citeauthoryear{Church and Hanks}{Church and Hanks}{1990}]%
        {church-hanks-1990-word}
\bibfield{author}{\bibinfo{person}{Kenneth~Ward Church} {and}
  \bibinfo{person}{Patrick Hanks}.} \bibinfo{year}{1990}\natexlab{}.
\newblock \showarticletitle{Word Association Norms, Mutual Information, and
  Lexicography}.
\newblock \bibinfo{journal}{\emph{Computational Linguistics}}
  \bibinfo{volume}{16}, \bibinfo{number}{1} (\bibinfo{year}{1990}),
  \bibinfo{pages}{22--29}.
\newblock
\urldef\tempurl%
\url{https://www.aclweb.org/anthology/J90-1003}
\showURL{%
\tempurl}


\bibitem[\protect\citeauthoryear{Deng, Dong, Socher, Li, Li, and Fei-Fei}{Deng
  et~al\mbox{.}}{2009}]%
        {imagenet_cvpr09}
\bibfield{author}{\bibinfo{person}{J. Deng}, \bibinfo{person}{W. Dong},
  \bibinfo{person}{R. Socher}, \bibinfo{person}{L.-J. Li}, \bibinfo{person}{K.
  Li}, {and} \bibinfo{person}{L. Fei-Fei}.} \bibinfo{year}{2009}\natexlab{}.
\newblock \showarticletitle{{ImageNet: A Large-Scale Hierarchical Image
  Database}}. In \bibinfo{booktitle}{\emph{CVPR09}}.
\newblock


\bibitem[\protect\citeauthoryear{Dwork, Hardt, Pitassi, Reingold, and
  Zemel}{Dwork et~al\mbox{.}}{2011}]%
        {DBLP:journals/corr/abs-1104-3913}
\bibfield{author}{\bibinfo{person}{Cynthia Dwork}, \bibinfo{person}{Moritz
  Hardt}, \bibinfo{person}{Toniann Pitassi}, \bibinfo{person}{Omer Reingold},
  {and} \bibinfo{person}{Richard~S. Zemel}.} \bibinfo{year}{2011}\natexlab{}.
\newblock \showarticletitle{Fairness Through Awareness}.
\newblock \bibinfo{journal}{\emph{CoRR}}  \bibinfo{volume}{abs/1104.3913}
  (\bibinfo{year}{2011}).
\newblock
\showeprint[arxiv]{1104.3913}
\urldef\tempurl%
\url{http://arxiv.org/abs/1104.3913}
\showURL{%
\tempurl}


\bibitem[\protect\citeauthoryear{Everingham, Eslami, {Van Gool}, Williams,
  Winn, and Zisserman}{Everingham et~al\mbox{.}}{2015}]%
        {PascalRetro}
\bibfield{author}{\bibinfo{person}{Mark Everingham}, \bibinfo{person}{{S. M.
  Ali} Eslami}, \bibinfo{person}{Luc {Van Gool}}, \bibinfo{person}{{Christopher
  K. I.} Williams}, \bibinfo{person}{John Winn}, {and} \bibinfo{person}{Andrew
  Zisserman}.} \bibinfo{year}{2015}\natexlab{}.
\newblock \showarticletitle{The Pascal Visual Object Classes Challenge: A
  Retrospective}.
\newblock \bibinfo{journal}{\emph{International Journal of Computer Vision}}
  \bibinfo{volume}{111}, \bibinfo{number}{1} (\bibinfo{date}{Jan.}
  \bibinfo{year}{2015}), \bibinfo{pages}{98--136}.
\newblock
\showISSN{0920-5691}
\urldef\tempurl%
\url{https://doi.org/10.1007/s11263-014-0733-5}
\showDOI{\tempurl}


\bibitem[\protect\citeauthoryear{Fano}{Fano}{1961}]%
        {fano1961transmission}
\bibfield{author}{\bibinfo{person}{Robert~M Fano}.}
  \bibinfo{year}{1961}\natexlab{}.
\newblock \showarticletitle{{Transmission of information: A statistical theory
  of communications}}.
\newblock \bibinfo{journal}{\emph{American Journal of Physics}}
  \bibinfo{volume}{29} (\bibinfo{year}{1961}), \bibinfo{pages}{793--794}.
\newblock


\bibitem[\protect\citeauthoryear{Greenwald, McGhee, and Schwartz}{Greenwald
  et~al\mbox{.}}{1998}]%
        {IAT}
\bibfield{author}{\bibinfo{person}{A.~G. Greenwald}, \bibinfo{person}{D.~E.
  McGhee}, {and} \bibinfo{person}{J.~L. Schwartz}.}
  \bibinfo{year}{1998}\natexlab{}.
\newblock \showarticletitle{Measuring individual differences in implicit
  cognition: the implicit association test}.
\newblock \bibinfo{journal}{\emph{Journal of personality and social
  psychology}}  \bibinfo{volume}{74} (\bibinfo{year}{1998}).
\newblock
Issue 6.


\bibitem[\protect\citeauthoryear{Hardt, Price, and Srebro}{Hardt
  et~al\mbox{.}}{2016}]%
        {hardt2016equality}
\bibfield{author}{\bibinfo{person}{Moritz Hardt}, \bibinfo{person}{Eric Price},
  {and} \bibinfo{person}{Nathan Srebro}.} \bibinfo{year}{2016}\natexlab{}.
\newblock \bibinfo{title}{Equality of Opportunity in Supervised Learning}.
\newblock
\newblock
\showeprint[arxiv]{1610.02413}~[cs.LG]


\bibitem[\protect\citeauthoryear{Harris}{Harris}{1954}]%
        {harris54}
\bibfield{author}{\bibinfo{person}{Zellig Harris}.}
  \bibinfo{year}{1954}\natexlab{}.
\newblock \showarticletitle{Distributional structure}.
\newblock \bibinfo{journal}{\emph{Word}} \bibinfo{volume}{10},
  \bibinfo{number}{2-3} (\bibinfo{year}{1954}), \bibinfo{pages}{146--162}.
\newblock
\urldef\tempurl%
\url{https://doi.org/10.1007/978-94-009-8467-7_1}
\showDOI{\tempurl}


\bibitem[\protect\citeauthoryear{Jurafsky and Martin}{Jurafsky and
  Martin}{2009}]%
        {Jurafsky2009}
\bibfield{author}{\bibinfo{person}{Dan Jurafsky} {and}
  \bibinfo{person}{James~H. Martin}.} \bibinfo{year}{2009}\natexlab{}.
\newblock \bibinfo{booktitle}{\emph{Speech and language processing : an
  introduction to natural language processing, computational linguistics, and
  speech recognition}}.
\newblock \bibinfo{publisher}{Pearson Prentice Hall}, \bibinfo{address}{Upper
  Saddle River, N.J.}
\newblock
\showISBNx{9780131873216 0131873210}
\urldef\tempurl%
\url{http://www.amazon.com/Speech-Language-Processing-2nd-Edition/dp/0131873210/ref=pd_bxgy_b_img_y}
\showURL{%
\tempurl}


\bibitem[\protect\citeauthoryear{Kendall}{Kendall}{1938}]%
        {kendall}
\bibfield{author}{\bibinfo{person}{M.~G. Kendall}.}
  \bibinfo{year}{1938}\natexlab{}.
\newblock \showarticletitle{{A New Measure of Rank Correlation}}.
\newblock \bibinfo{journal}{\emph{Biometrika}} \bibinfo{volume}{30},
  \bibinfo{number}{1-2} (\bibinfo{date}{06} \bibinfo{year}{1938}),
  \bibinfo{pages}{81--93}.
\newblock
\showeprint{https://academic.oup.com/biomet/article-pdf/30/1-2/81/423380/30-1-2-81.pdf}
\urldef\tempurl%
\url{https://doi.org/10.1093/biomet/30.1-2.81}
\showURL{%
\tempurl}


\bibitem[\protect\citeauthoryear{Kuznetsova, Rom, Alldrin, Uijlings, Krasin,
  Pont{-}Tuset, Kamali, Popov, Malloci, Duerig, and Ferrari}{Kuznetsova
  et~al\mbox{.}}{2018}]%
        {OpenImages}
\bibfield{author}{\bibinfo{person}{Alina Kuznetsova}, \bibinfo{person}{Hassan
  Rom}, \bibinfo{person}{Neil Alldrin}, \bibinfo{person}{Jasper R.~R.
  Uijlings}, \bibinfo{person}{Ivan Krasin}, \bibinfo{person}{Jordi
  Pont{-}Tuset}, \bibinfo{person}{Shahab Kamali}, \bibinfo{person}{Stefan
  Popov}, \bibinfo{person}{Matteo Malloci}, \bibinfo{person}{Tom Duerig}, {and}
  \bibinfo{person}{Vittorio Ferrari}.} \bibinfo{year}{2018}\natexlab{}.
\newblock \showarticletitle{The Open Images Dataset {V4:} Unified image
  classification, object detection, and visual relationship detection at
  scale}.
\newblock \bibinfo{journal}{\emph{CoRR}}  \bibinfo{volume}{abs/1811.00982}
  (\bibinfo{year}{2018}).
\newblock
\showeprint[arxiv]{1811.00982}
\urldef\tempurl%
\url{http://arxiv.org/abs/1811.00982}
\showURL{%
\tempurl}


\bibitem[\protect\citeauthoryear{Lin, Maire, Belongie, Bourdev, Girshick, Hays,
  Perona, Ramanan, Doll{\'{a}}r, and Zitnick}{Lin et~al\mbox{.}}{2014}]%
        {COCO}
\bibfield{author}{\bibinfo{person}{Tsung{-}Yi Lin}, \bibinfo{person}{Michael
  Maire}, \bibinfo{person}{Serge~J. Belongie}, \bibinfo{person}{Lubomir~D.
  Bourdev}, \bibinfo{person}{Ross~B. Girshick}, \bibinfo{person}{James Hays},
  \bibinfo{person}{Pietro Perona}, \bibinfo{person}{Deva Ramanan},
  \bibinfo{person}{Piotr Doll{\'{a}}r}, {and} \bibinfo{person}{C.~Lawrence
  Zitnick}.} \bibinfo{year}{2014}\natexlab{}.
\newblock \showarticletitle{Microsoft {COCO:} Common Objects in Context}.
\newblock \bibinfo{journal}{\emph{CoRR}}  \bibinfo{volume}{abs/1405.0312}
  (\bibinfo{year}{2014}).
\newblock
\showeprint[arxiv]{1405.0312}
\urldef\tempurl%
\url{http://arxiv.org/abs/1405.0312}
\showURL{%
\tempurl}


\bibitem[\protect\citeauthoryear{Role and Nadif.}{Role and Nadif.}{2011}]%
        {kdir11}
\bibfield{author}{\bibinfo{person}{Fran\c{C}ois Role} {and}
  \bibinfo{person}{Mohamed Nadif.}} \bibinfo{year}{2011}\natexlab{}.
\newblock \showarticletitle{Handling the Impact of Low Frequency Events on
  Co-Occurrence Based Measures of Word Similarity - A Case Study of Pointwise
  Mutual Information}. In \bibinfo{booktitle}{\emph{Proceedings of the
  International Conference on Knowledge Discovery and Information Retrieval -
  KDIR, (IC3K 2011)}}. \bibinfo{publisher}{SciTePress},
  \bibinfo{pages}{218--223}.
\newblock


\bibitem[\protect\citeauthoryear{Russakovsky, Deng, Su, Krause, Satheesh, Ma,
  Huang, Karpathy, Khosla, Bernstein, Berg, and Fei-Fei}{Russakovsky
  et~al\mbox{.}}{2015}]%
        {ILSVRC15}
\bibfield{author}{\bibinfo{person}{Olga Russakovsky}, \bibinfo{person}{Jia
  Deng}, \bibinfo{person}{Hao Su}, \bibinfo{person}{Jonathan Krause},
  \bibinfo{person}{Sanjeev Satheesh}, \bibinfo{person}{Sean Ma},
  \bibinfo{person}{Zhiheng Huang}, \bibinfo{person}{Andrej Karpathy},
  \bibinfo{person}{Aditya Khosla}, \bibinfo{person}{Michael Bernstein},
  \bibinfo{person}{Alexander~C. Berg}, {and} \bibinfo{person}{Li Fei-Fei}.}
  \bibinfo{year}{2015}\natexlab{}.
\newblock \showarticletitle{{ImageNet Large Scale Visual Recognition
  Challenge}}.
\newblock \bibinfo{journal}{\emph{International Journal of Computer Vision
  (IJCV)}} \bibinfo{volume}{115}, \bibinfo{number}{3} (\bibinfo{year}{2015}),
  \bibinfo{pages}{211--252}.
\newblock
\urldef\tempurl%
\url{https://doi.org/10.1007/s11263-015-0816-y}
\showDOI{\tempurl}


\bibitem[\protect\citeauthoryear{Shannon}{Shannon}{1948}]%
        {journals/bstj/Shannon48}
\bibfield{author}{\bibinfo{person}{Claude~E. Shannon}.}
  \bibinfo{year}{1948}\natexlab{}.
\newblock \showarticletitle{A mathematical theory of communication.}
\newblock \bibinfo{journal}{\emph{Bell Syst. Tech. J.}} \bibinfo{volume}{27},
  \bibinfo{number}{3} (\bibinfo{year}{1948}), \bibinfo{pages}{379--423}.
\newblock
\urldef\tempurl%
\url{http://dblp.uni-trier.de/db/journals/bstj/bstj27.html#Shannon48}
\showURL{%
\tempurl}


\bibitem[\protect\citeauthoryear{Snow}{Snow}{2018}]%
        {ACLURekognition}
\bibfield{author}{\bibinfo{person}{Jacob Snow}.}
  \bibinfo{year}{2018}\natexlab{}.
\newblock \showarticletitle{Amazon’s Face Recognition Falsely Matched 28
  Members of Congress With Mugshots}.
\newblock  (\bibinfo{year}{2018}).
\newblock


\bibitem[\protect\citeauthoryear{Spearman}{Spearman}{1904}]%
        {spearman04}
\bibfield{author}{\bibinfo{person}{C. Spearman}.}
  \bibinfo{year}{1904}\natexlab{}.
\newblock \showarticletitle{The Proof and Measurement of Association Between
  Two Things}.
\newblock \bibinfo{journal}{\emph{American Journal of Psychology}}
  \bibinfo{volume}{15} (\bibinfo{year}{1904}), \bibinfo{pages}{88--103}.
\newblock


\bibitem[\protect\citeauthoryear{{Stanford Vision Lab}}{{Stanford Vision
  Lab}}{2020}]%
        {ImageNet}
\bibfield{author}{\bibinfo{person}{{Stanford Vision Lab}}.}
  \bibinfo{year}{2020}\natexlab{}.
\newblock \showarticletitle{ImageNet}.
\newblock \bibinfo{journal}{\emph{http://image-net.org/explore}}
  (\bibinfo{year}{2020}).
\newblock
\newblock
\shownote{accessed 6.Oct.2020.}


\bibitem[\protect\citeauthoryear{Stock and Cisse}{Stock and Cisse}{2018}]%
        {stock2018convnets}
\bibfield{author}{\bibinfo{person}{Pierre Stock} {and}
  \bibinfo{person}{Moustapha Cisse}.} \bibinfo{year}{2018}\natexlab{}.
\newblock \showarticletitle{Convnets and imagenet beyond accuracy:
  Understanding mistakes and uncovering biases}. In
  \bibinfo{booktitle}{\emph{Proceedings of the European Conference on Computer
  Vision (ECCV)}}. \bibinfo{pages}{498--512}.
\newblock


\bibitem[\protect\citeauthoryear{Toglia and Battig}{Toglia and Battig}{1978}]%
        {WordNorms}
\bibfield{author}{\bibinfo{person}{M.~P. Toglia} {and} \bibinfo{person}{W.~F.
  Battig}.} \bibinfo{year}{1978}\natexlab{}.
\newblock \bibinfo{booktitle}{\emph{Handbook of semantic word norms.}}
\newblock \bibinfo{publisher}{Lawrence Erlbaum}.
\newblock


\bibitem[\protect\citeauthoryear{Wilson, Hoffman, and Morgenstern}{Wilson
  et~al\mbox{.}}{2019}]%
        {DBLP:journals/corr/abs-1902-11097}
\bibfield{author}{\bibinfo{person}{Benjamin Wilson}, \bibinfo{person}{Judy
  Hoffman}, {and} \bibinfo{person}{Jamie Morgenstern}.}
  \bibinfo{year}{2019}\natexlab{}.
\newblock \showarticletitle{Predictive Inequity in Object Detection}.
\newblock \bibinfo{journal}{\emph{CoRR}}  \bibinfo{volume}{abs/1902.11097}
  (\bibinfo{year}{2019}).
\newblock
\showeprint[arxiv]{1902.11097}
\urldef\tempurl%
\url{http://arxiv.org/abs/1902.11097}
\showURL{%
\tempurl}


\end{thebibliography}
\pagebreak
\appendix 


\onecolumn
\setcounter{secnumdepth}{3}
\section*{Appendix}

The following sections break down all of the different metrics we examined.  In Section \ref{app:gap_calcs}, Gap Calculations, we present the calculations of the label association metric {\it gaps}, a measurement of the skew with respect to the given sensitive labels $x_1$ and $x_2$, such as $woman$ $(x_1)$ and $man$ $(x_2)$. In Section \ref{app:orientation_details}, Metric Orientations, we present the derivations to calculate their {\it orientations}, which represent the theoretical sensitivities of each metric to various marginal or joint label probabilities.

\section{Gap Calculations}\label{app:gap_calcs}

\begin{itemize}
\item Demographic Parity ($DP$):
\[
DP\_gap = p(y|x_1) - p(y|x_2)
\]
\item Sørensen-Dice Coefficient ($SDC$):
\[
SDC\_gap=
\frac{p(x_1,y)}{p(x_1)+p(y)}
-\frac{p(x_2,y)}{p(x_2)+p(y)}
\]
\item Jaccard Index ($JI$):
\[
JI\_gap=
\frac{p(x_1,y)}{p(x_1)+p(y)-p(x_1,y)}
-\frac{p(x_2,y)}{p(x_2)+p(y)-p(x_2,y)}
\]
\item Log-Likelihood Ratio ($LLR$):
\[
LLR\_gap = ln(p(x_1|y)) - ln(p(x_2|y))
\]
\item Pointwise Mutual Information ($PMI$):
\[
PMI\_gap=ln\left(\frac{p(x_1,y)}{p(x_1)p(y)}\right) -
ln\left(\frac{p(x_2,y)}{p(x_2)p(y)}\right)
\]
\item Normalized Pointwise Mutual Information, p(y) normalization ($nPMI_y$):
\[
nPMI_y\_gap=\frac
{ln\left(\frac{p(x_1,y)}{p(x_1)p(y)}\right)}
{ln\left(p(y)\right)} -
\frac
{ln\left(\frac{p(x_2,y)}{p(x_2)p(y)}\right)}
{ln\left(p(y)\right)}
\]
\item Normalized Pointwise Mutual Information, p(x,y) normalization ($nPMI_{xy}$):
\[
nPMI_{xy}\_gap=\frac
{ln\left(\frac{p(x_1,y)}{p(x_1)p(y)}\right)}
{ln\left(p(x_1,y)\right)} -
\frac
{ln\left(\frac{p(x_2,y)}{p(x_2)p(y)}\right)}
{ln\left(p(x_2,y)\right)}
\]
\item Squared Pointwise Mutual Information ($PMI^2$):
\[
PMI^2\_gap=ln\left(\frac{p(x_1,y)^2}{p(x_1)p(y)}\right) -
ln\left(\frac{p(x_2,y)^2}{p(x_2)p(y)}\right)
\]
\item Kendall Rank Correlation ($\tau_b$):
\[
\tau_b = \frac{n_c-n_d}{\sqrt{(n_0-n_1)(n_0-n_2)}}
\]
\[
\tau_b\_gap=\tau_b(l_1=x_1,l_2=y)-\tau_b(l_1=x_2,l_2=y)
\]
\item t-test:
\[
t \mhyphen test\_gap=\frac{p(x_1,y)-p(x_1)p(y)}{\sqrt{p(x_1)p(y)}}
- \frac{p(x_2,y)-p(x_2)p(y)}{\sqrt{p(x_2)p(y)}}
\]
\end{itemize}

\section{Mathematical Reductions}

\begin{itemize}
\item{PMI:}
\begin{align}
\begin{split}
PMI\_gap=ln\left(\frac{p(x_1,y)}{p(x_1)p(y)}\right) -
ln\left(\frac{p(x_2,y)}{p(x_2)p(y)}\right)
\end{split} \\
\begin{split}
PMI\_gap=ln\left(
\frac{p(x_1,y)}{p(x_1)p(y)}.
\frac{p(x_2)p(y)}{p(x_2,y)}\right)
\end{split} \\
\begin{split}
PMI\_gap=ln\left(
\frac{p(x_1,y)}{p(x_1)}.
\frac{p(x_2)}{p(x_2,y)}\right)
\end{split} \\
\begin{split}
PMI\_gap=ln\left(
\frac{p(y|x_1)}{p(y|x_2)}\right)
\end{split} \\
\begin{split}
PMI\_gap=
ln(p(y|x_1))-ln(p(y|x_2))
\end{split}
\end{align}

\item{nPMI p(y) normalization:}
\begin{align}
\begin{split}
nPMI\_gap=\frac
{ln\left(\frac{p(x_1,y)}{p(x_1)p(y)}\right)}
{ln\left(p(y)\right)} -
\frac
{ln\left(\frac{p(x_2,y)}{p(x_2)p(y)}\right)}
{ln\left(p(y)\right)}
\end{split} \\
\begin{split}
nPMI\_gap=
\frac{ln(p(y|x_1))-ln(p(y|x_2))}{ln(p(y))}
\end{split}
\end{align}

\item{nPMI p(x,y) normalization:}
\begin{align}
\begin{split}
nPMI\_gap=\frac
{ln\left(\frac{p(x_1,y)}{p(x_1)p(y)}\right)}
{ln\left(p(x_1,y)\right)} -
\frac
{ln\left(\frac{p(x_2,y)}{p(x_2)p(y)}\right)}
{ln\left(p(x_2,y)\right)}
\end{split} \\
\begin{split}
nPMI\_gap=
ln\left({\left(\frac{p(x_1,y)}{p(x_1)p(y)}\right)}^{1/ln(p(x_1,y))}\right) -
ln\left({\left(\frac{p(x_2,y)}{p(x_2)p(y)}\right)}^{1/ln(p(x_2,y))}\right)
\end{split} \\
\begin{split}
nPMI\_gap=
ln\left(
\frac
{{\left(\frac{p(x_1,y)}{p(x_1)p(y)}\right)}^{1/ln(p(x_1,y))}}
{{\left(\frac{p(x_2,y)}{p(x_2)p(y)}\right)^{1/ln(p(x_2,y))}}}
\right)
\end{split} \\
\begin{split}
nPMI\_gap=
ln\left(
\frac
{{\left(\frac{p(x_1,y)}{p(x_1)}\right)}^{1/ln(p(x_1,y))}}
{{\left(\frac{p(x_2,y)}{p(x_2)}\right)^{1/ln(p(x_2,y))}}}
.{p(y)}^{(1/ln(p(x_1,y))-1/ln(p(x_2,y)))}
\right)
\end{split} \\
\begin{split}
nPMI\_gap=
ln\left(
\frac
{{\left(\frac{p(x_1,y)}{p(x_1)}\right)}^{1/ln(p(x_1,y))}}
{{\left(\frac{p(x_2,y)}{p(x_2)}\right)^{1/ln(p(x_2,y))}}}
\right)
+ln\left(
{p(y)}^{(1/ln(p(x_1,y))-1/ln(p(x_2,y)))}
\right)
\end{split} \\
\begin{split}
nPMI\_gap=
ln\left(
\frac
{{p(y|x_1)}^{1/ln(p(x_1,y))}}
{{p(y|x_2)^{1/ln(p(x_2,y))}}}
\right)
+ ln(p(y)).\left(\frac{1}{ln(p(x_1,y))}-\frac{1}{ln(p(x_2,y))}\right)
\end{split}\\
\begin{split}
nPMI\_gap=
\frac{ln(p(y|x_1))}{ln(p(x_1,y))}
- \frac{ln(p(y|x_2))}{ln(p(x_2,y))}
+ ln(p(y)).\left(\frac{ln(\frac{p(x_2,y)}{p(x_1,y)})}{ln(p(x_1,y))ln(p(x_2,y))}\right)
\end{split} \\
\begin{split}
nPMI\_gap=
\frac{ln(p(x_2))}{ln(p(x_2,y))}
- \frac{ln(p(x_1))}{ln(p(x_1,y))}
+ \frac{ln(p(y))ln(\frac{p(x_2,y)}{p(x_1,y)})}{ln(p(x_1,y))ln(p(x_2,y))}
\end{split} \\
\begin{split}
nPMI\_gap=
\frac{ln(p(x_2))}{ln(p(x_2,y))}
- \frac{ln(p(x_1))}{ln(p(x_1,y))}
+ \frac{ln(p(y))}{ln(p(x_1,y))}
- \frac{ln(p(y))}{ln(p(x_2,y))}
\end{split} \\
\end{align}
\item{PMI$^2$:}
\begin{align}
\begin{split}
PMI^2\_gap=ln\left(\frac{p(x_1,y)^2}{p(x_1)p(y)}\right) -
ln\left(\frac{p(x_2,y)^2}{p(x_2)p(y)}\right)
\end{split} \\
\begin{split}
PMI^2\_gap=ln\left(
\frac{p(x_1,y)^2}{p(x_1)p(y)}.
\frac{p(x_2)p(y)}{p(x_2,y)^2}\right)
\end{split} \\
\begin{split}
PMI^2\_gap=ln\left(
\frac{p(x_1,y)^2}{p(x_1)}.
\frac{p(x_2)}{p(x_2,y)^2}\right)
\end{split} \\
\begin{split}
PMI^2\_gap=ln\left(
\frac{p(y|x_1)p(x_1,y)}{p(y|x_2)p(x_2,y)}\right)
\end{split} \\
\begin{split}
PMI^2\_gap=ln(p(y|x_1))-ln(p(y|x_2))+ln(p(x_1,y))-ln(p(x_2,y))
\end{split}
\end{align}

\item{Kendall Rank Correlation ($\tau_b$):} \\

\noindent Formal definition:
\[
\tau_b = \frac{n_c-n_d}{\sqrt{(n_0-n_1)(n_0-n_2)}}
\]
\begin{center}
\begin{itemize}
    \item $n_0=n(n-1)/2$
    \item $n_1=\sum_{i}t_i(t_i-1)/2$
    \item $n_2=\sum_{j}u_j(u_j-1)/2$
    \item $n_c$ is number of concordant pairs
    \item $n_d$ is number of discordant pairs
    \item $n_c$ is number of concordant pairs
    \item $t_i$ is number of tied values in the $i^{th}$ group of ties for the first quantity
    \item $u_j$ is number of tied values in the $j^{th}$ group of ties for the second quantity
\end{itemize}
\end{center}

New notations for our use case:
\[
\tau_b(l_1=x_1,l_2=y) = \frac{n_c(l_1=x_1,l_2=y)-n_d(l_1=x_1,l_2=y)}{\sqrt{({n \choose 2}-n_s(l=x_1))({n\choose 2}-n_s(l=y))}}
\]
\begin{itemize}
    \item $n_c(l_1,l_2)={C_{l_1=0,l_2=0}\choose 2}+{C_{l_1=1,l_2=1}\choose 2}$
    \item $n_d(l_1,l_2)={C_{l_1=0,l_2=1}\choose 2}+{C_{l_1=1,l_2=0}\choose 2}$
    \item $n_s(l)={C_{l=0}\choose 2}+{C_{l=1}\choose 2}$
    \item $C_{conditions}$ is number of examples which satisfies the conditions
    \item $n$ number of examples in data set
\end{itemize}

\begin{align}
\begin{split}
\tau_b(l_1=x,l_2=y)=\frac{n^2(1-2p(x)-2p(y)+2p(x,y)+2p(x)p(y))+n(2p(x)+2p(y)-4p(x,y)-1)}{n^2\sqrt{(p(x)-p(x)^2)(p(y)-p(y)^2)}}
\end{split} \\
\begin{split}
\tau_b\_gap=\tau_b(l_1=x_1,l_2=y)-\tau_b(l_1=x_2,l_2=y)
\end{split}
\end{align}
\end{itemize}

\section{Metric Orientations}\label{app:orientation_details}

In this section, we derive the orientations of each metric, which can be thought of as the sensitivity of each metric to the given label probability.  Table \ref{tab:orientation_table} provides a summary, followed by the longer forms.

\begin{table}[H]
\caption{Metric Orientations}\label{tab:orientation_table}
\begin{tabular}{ |c|c|c|c|  }
\hline
&$\partial p(y)$&$\partial p(x_1, y)$&$\partial p(x_2, y)$\\
\hline
&&&\\[0.2ex]
$\partial$DP&$0$&$\frac{1}{p(x_1)}$&$\frac{-1}{p(x_2)}$\\[2.5ex]
$\partial$PMI&$0$&$\frac{1}{p(x_1,y)}$&$\frac{-1}{p(x_2,y)}$\\[2.5ex]
$\partial nPMI_{y}$&$\frac{ln(\frac{p(x_2|y)}{p(x_1|y)})}{ln^2(p(y))p(y)}$&$\frac{1}{ln(p(y))p(x_1, y)}$&$\frac{-1}{ln(p(y))p(x_2, y)}$\\[2.5ex]
$\partial nPMI_{xy}$&$\frac{1}{ln(p(x_1,y))p(y)}-\frac{1}{ln(p(x_2,y))p(y)}$&$\frac{ln(p(y))-ln(p(x_1))}{ln^2(p(x_1,y))p(x_1,y)}$&$\frac{ln(p(x_2))-ln(p(y))}{ln^2(p(x_2,y))p(x_2,y)}$\\[2.5ex]
$\partial PMI^2$&0&$\frac{2}{p(x_1,y)}$&$\frac{-2}{p(x_2,y)}$\\[2.5ex]
$\partial$SDC&$\frac{p(x_2,y)}{(p(x_2)+p(y))^2}-\frac{p(x_1,y)}{(p(x_1)+p(y))^2}$&$\frac{1}{p(x_1)+p(y)}$&$\frac{-1}{p(x_2)+p(y)}$\\[2.5ex]
$\partial$JI&$\frac{p(x_1,y)}{p(x_1)+p(y)-p(x_1,y)}
-\frac{p(x_2,y)}{p(x_2)+p(y)-p(x_2,y)}$&$\frac{p(x_1)+p(y)}{(p(x_1)+p(y)-p(x_1,y))^2}$&$\frac{p(x_2)+p(y)}{(p(x_2)+p(y)-p(x_2,y))^2}$\\[2.5ex]
$\partial$LLR&$0$&$\frac{1}{p(x_1, y)}$&$\frac{-1}{p(x_2, y)}$\\[2.5ex]
$\partial \tau_b$&\textit{annoyingly long}&$\frac{(2-\frac{4}{n})}{\sqrt{(p(x_1)-p(x_1)^2)(p(y)-p(y)^2)}}$&$\frac{(\frac{4}{n}-2)}{\sqrt{(p(x_2)-p(x_2)^2)(p(y)-p(y)^2)}}$\\[2.5ex]
$\partial t \mhyphen test\_gap$&$\frac{\sqrt{p(x_2)}-\sqrt{p(x_1)}}{2\sqrt{p(y)}}$&$\frac{1}{\sqrt{p(x_1)*p(y)}}$&$\frac{-1}{\sqrt{p(x_2)*p(y)}}$\\[2.5ex]
\hline
\end{tabular}
\end{table}

\begin{itemize}
    \item {Demographic Parity:}
\begin{align}
\begin{split}
    DP\_gap ={}& p(y|x_1) - p(y|x_2) 
\end{split}\\
\begin{split}
\frac{\partial DP\_gap}{\partial p(y)} ={}& 0
\end{split}\\
\begin{split}
\frac{\partial DP\_gap}{\partial p(x_1, y)} ={}&  \frac{1}{p(x_1)}
\end{split}\\
\begin{split}
\frac{\partial DP\_gap}{\partial p(x_2, y)} ={}&  \frac{-1}{p(x_2)}
\end{split}
\end{align}

\item{Sørensen-Dice Coefficient:}
\begin{align}
\begin{split}
SDC\_gap=\frac{p(x_1,y)}{p(x_1)+p(y)}-\frac{p(x_2,y)}{p(x_2)+p(y)}
\end{split} \\
\begin{split}
\frac{\partial SDC\_gap}{\partial p(y)}=\frac{p(x_2,y)}{(p(x_2)+p(y))^2}-\frac{p(x_1,y)}{(p(x_1)+p(y))^2}
\end{split} \\
\begin{split}
\frac{\partial SDC\_gap}{\partial p(x_1, y)}=\frac{1}{p(x_1)+p(y)}
\end{split} \\
\begin{split}
\frac{\partial SDC\_gap}{\partial p(x_2, y)}=\frac{-1}{p(x_2)+p(y)}
\end{split} \\
\end{align}

\item{Jaccard Index:}
\begin{align}
\begin{split}
JI\_gap=
\frac{p(x_1,y)}{p(x_1)+p(y)-p(x_1,y)}
-\frac{p(x_2,y)}{p(x_2)+p(y)-p(x_2,y)}
\end{split} \\
\begin{split}
\frac{\partial JI\_gap}{\partial p(y)}=
\frac{p(x_2,y)}{(p(x_2)+p(y)-p(x_2,y))^2}
-\frac{p(x_1,y)}{(p(x_1)+p(y)-p(x_1,y))^2}
\end{split} \\
\begin{split}
\frac{\partial JI\_gap}{\partial p(x_1,y)}=
\frac{p(x_1)+p(y)}{(p(x_1)+p(y)-p(x_1,y))^2}
\end{split} \\
\begin{split}
\frac{\partial JI\_gap}{\partial p(x_2,y)}=
\frac{p(x_2)+p(y)}{(p(x_2)+p(y)-p(x_2,y))^2}
\end{split} \\
\end{align}

\item{Log-Likelihood Ratio:}
\begin{align}
\begin{split}
LLR\_gap = ln(p(x_1|y)) - ln(p(x_2|y))
\end{split} \\
\begin{split}
\frac{\partial LLR\_gap}{\partial p(y)} = \frac{p(x_2,y)}{p(x_2|y)p(y)^2} - \frac{p(x_1,y)}{p(x_1|y)p(y)^2} = 0
\end{split} \\
\begin{split}
\frac{\partial LLR\_gap}{\partial p(x_1, y)} = \frac{1}{p(x_1|y)p(y)} = \frac{1}{p(x_1, y)}
\end{split} \\
\begin{split}
\frac{\partial LLR\_gap}{\partial p(x_2, y)} = \frac{-1}{p(x_2, y)}
\end{split} \\
\end{align}

\item{PMI:}
\begin{align}
\begin{split}
PMI\_gap=
ln(p(y|x_1))-ln(p(y|x_2))
\end{split} \\
\begin{split}
\frac{\partial PMI\_gap}{\partial p(y)} = 0
\end{split} \\
\begin{split}
\frac{\partial PMI\_gap}{\partial p(x_1, y)} = \frac{1}{p(x_1,y)}
\end{split} \\
\begin{split}
\frac{\partial PMI\_gap}{\partial p(x_2, y)} = \frac{1}{p(x_2,y)}
\end{split}
\end{align}

\item{PMI$^2$:}
\begin{align}
\begin{split}
PMI^2\_gap=ln(p(y|x_1))-ln(p(y|x_2))+ln(p(x_1,y))-ln(p(x_2,y))
\end{split} \\
\begin{split}
\frac{\partial PMI^2\_gap}{\partial p(y)} = 0
\end{split} \\
\begin{split}
\frac{\partial PMI^2\_gap}{\partial p(x_1, y)} = \frac{2}{p(x_1,y)}
\end{split} \\
\begin{split}
\frac{\partial PMI^2\_gap}{\partial p(x_2, y)} = \frac{-2}{p(x_2,y)}
\end{split}
\end{align}

\item{nPMI, normalized by p(y) :}
\begin{align}
\begin{split}
nPMI\_gap = \frac{ln(p(y|x_1))}{ln(p(y))} - \frac{ln(p(y|x_2))}{ln(p(y))}
\end{split} \\
\begin{split}
\frac{\partial nPMI\_gap}{\partial p(y)} = \frac{ln(p(y|x_2))}{ln^2(p(y))p(y)} - \frac{ln(p(y|x_1))}{ln^2(p(y))p(y)} = \frac{ln(\frac{p(x_2|y)}{p(x_1|y)})}{ln^2(p(y))p(y)}
\end{split} \\
\begin{split}
\frac{\partial nPMI\_gap}{\partial p(x_1,y)} = \frac{1}{ln(p(y))p(x_1, y)}
\end{split} \\
\begin{split}
\frac{\partial nPMI\_gap}{\partial p(x_2,y)} = \frac{-1}{ln(p(y))p(x_2, y)}
\end{split}
\end{align}

\item{nPMI, normalized by p(x,y) :}
\begin{align}
\begin{split}\label{npmi_xy}
nPMI\_gap=
\frac{ln(p(x_2))}{ln(p(x_2,y))}
- \frac{ln(p(x_1))}{ln(p(x_1,y))}
+ \frac{ln(p(y))}{ln(p(x_1,y))}
- \frac{ln(p(y))}{ln(p(x_2,y))}
\end{split} \\
\begin{split}
\frac{\partial nPMI\_gap}{\partial p(y)} = \frac{1}{ln(p(x_1,y))p(y)}
- \frac{1}{ln(p(x_2,y))p(y)}
\end{split} \\
\begin{split}
\frac{\partial nPMI\_gap}{\partial p(x_1,y)} = \frac{ln(p(y))-ln(p(x_1))}{ln^2(p(x_1,y))p(x_1,y)}
\end{split} \\
\begin{split}
\frac{\partial nPMI\_gap}{\partial p(x_1,y)} = \frac{ln(p(x_2))-ln(p(y))}{ln^2(p(x_2,y))p(x_2,y)}
\end{split} \\
\end{align}

\item{Kendall rank correlation ($\tau_b$):}\\
\begin{align}
\begin{split}
\tau_b\_gap=\tau_b(l_1=x_1,l_2=y)-\tau_b(l_1=x_2,l_2=y)
\end{split} \\
\begin{split}
\frac{\partial \tau_b\_gap}{\partial p(x_1,y)}=\frac{(2-\frac{4}{n})}{\sqrt{(p(x_1)-p(x_1)^2)(p(y)-p(y)^2)}}
\end{split} \\
\begin{split}
\frac{\partial \tau_b\_gap}{\partial p(x_2,y)}=\frac{(2-\frac{4}{n})}{\sqrt{(p(x_2)-p(x_2)^2)(p(y)-p(y)^2)}}
\end{split} \\
\end{align}

\item{$t \mhyphen test$:}\\
\begin{align}
\begin{split}\label{ttest_gap}
t \mhyphen test\_gap=\frac{p(x_1,y)-p(x_1)*p(y)}{\sqrt{p(x_1)*p(y)}}
- \frac{p(x_2,y)-p(x_2)*p(y)}{\sqrt{p(x_2)*p(y)}}
\end{split} \\
\begin{split}
\frac{\partial t \mhyphen test\_gap}{\partial p(y)} = \frac{\sqrt{p(x_2)}-\sqrt{p(x_1)}}{2\sqrt{p(y)}}
\end{split} \\
\begin{split}
\frac{\partial t \mhyphen test\_gap}{\partial p(x_1,y)} = \frac{1}{\sqrt{p(x_1)*p(y)}}
\end{split} \\
\begin{split}
\frac{\partial t \mhyphen test\_gap}{\partial p(x_2,y)} = \frac{-1}{\sqrt{p(x_2)*p(y)}}
\end{split} \\
\end{align}
\end{itemize}

\section{Comparison Tables}
\begin{table}[H]
\caption{Mean/Std of top 100 Male Association Gaps}
\centering
\begin{tabular}{|c||c|c|}
\hline
\multicolumn{3}{|c|}{Scales of top 100 ($x_1=man$)} \\
\hline
Metrics&Mean/Std $C(y)$&Mean/Std $C(x_1,y)$\\
\hline
$DP$&$101.38 (\pm199.32)$&$1.00 (\pm0.00)$\\
$nPMI_{xy}$&$24371.15 (\pm27597.10)$&$237.92 (\pm604.29)$\\
$nPMI_{y}$&$75160.45(\pm119779.13)$&$5680.44 (\pm17193.19)$\\
$PMI$&$1875.86 (\pm3344.23)$&$2.43 (\pm4.72)$\\
$PMI^2$&$831.43 (\pm620.02)$&$1.09 (\pm0.32)$\\
$SDC$&$676.33 (\pm412.23)$&$1.00 (\pm0.00)$\\
$\tau_b$&$189294.69 (\pm149505.10)$&$24524.89 (\pm32673.77)$\\
\hline
\end{tabular}
\end{table}

\begin{table}[H]
\caption{Mean/Std of top 100 Male-Female Association Gaps}
\centering
\begin{tabular}{|c||l|l|l|}
\hline
\multicolumn{4}{|c|}{Counts for top 100 gaps($x_1$=MALE, $x_2$=FEMALE)} \\
\hline
Metrics&Mean/Std $C(y)$&Mean/Std $C(x_1,y)$&Mean/Std $C(x_2, y)$\\
\hline
$DP$&154960.81($\pm$147823.81)& 71552.62 ($\pm$67400.44)& 70398.07($\pm$56824.32)\\
$JI$& 70403.46 ($\pm$86766.14)& 29670.94 ($\pm$35381.09)& 35723.13($\pm$39308.46)\\
$LLR$&  1020.60 ($\pm$1971.77)&    57.95 ($\pm$178.59)&   558.36 ($\pm$1420.15)\\
$nPMI_{xy}$& 13869.42 ($\pm$46244.17)&  5771.73 ($\pm$23564.63)&  9664.46 ($\pm$31475.19)\\
$nPMI_y$& 23156.11 ($\pm$73662.90)&  6632.79 ($\pm$25631.26)& 10376.39 ($\pm$33559.27)\\
$PMI$&  1020.60 ($\pm$1971.77)&    57.95 ($\pm$178.59)&   558.36 ($\pm$1420.15)\\
$PMI^2$&  1020.60 ($\pm$1971.77)&    57.95 ($\pm$178.59)&   558.36 ($\pm$1420.15)\\
$SDC$& 67648.20 ($\pm$87258.41)& 28193.83 ($\pm$35295.25)& 34443.22 ($\pm$39501.21)\\
$\tau_b$&134076.65 ($\pm$144354.70)& 53845.80 ($\pm$58016.12)& 56032.09 ($\pm$52124.92)\\
\hline
\end{tabular}
\end{table}


\begin{table}[H]
\caption{Minimum/Maximum of top 100 Male-Female Association Gaps}\label{app_minmaxcounts}
\centering
\begin{tabular}{ |c|c|c|c|  }
\hline
Metrics&Min/Max $C(y)$&Min/Max $C(x_1,y)$&Min/Max $C(x_2, y)$\\
\hline
&&&\\
$PMI$&       15 / 10,551    &        1 / 1,059     &        8 / 7,755     \\[2ex]
$PMI^2$&       15 / 10,551    &        1 / 1,059     &        8 / 7,755     \\[2ex]
$LLR$&       15 / 10,551    &        1 / 1,059     &        8 / 7,755     \\[2ex]
$DP$&     6,104 / 785,045   &      628 / 239,950   &     5,347 / 197,795   \\[2ex]
$JI$&     4,158 / 562,445   &      399 / 144,185   &     3,359 / 183,132   \\[2ex]
$SDC$&     2,906 / 562,445   &     139 / 144,185  &     2,563 / 183,132   \\[2ex]
$nPMI_{y}$&       35 / 562,445   &       1 / 144,185  &        9 / 183,132   \\[2ex]
$nPMI_{xy}$&       34 / 270,748   &        1 / 144,185   &       20 / 183,132   \\[2ex]
$\tau_b$&     6,104 / 785,045   &      628 / 207,723   &     5,347 / 183,132   \\[2ex]
$t \mhyphen test$&      960 / 562,445   &       72 / 144,185   &      870 / 183,132   \\[2ex]
\hline
\end{tabular}%
\end{table}

\newpage
\section{Comparison Plots}
\subsection{Overall Rank Changes}

\begin{table}[htb]
\caption{Demographic Parity (di), Jaccard Index (ji), Sørensen-Dice Coefficient (sdc) Comparison}
\centering
\begin{adjustbox}{width=1.1\textwidth,center}
\begin{tabular}{cc}
\includegraphics{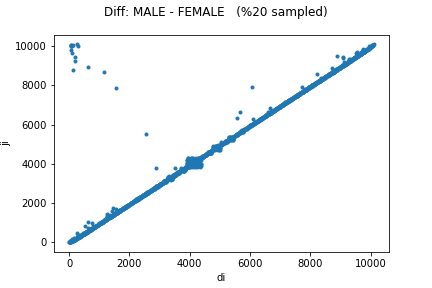} &
\includegraphics{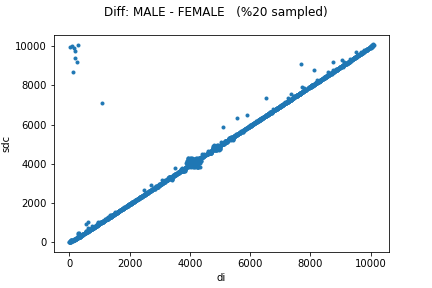} \\
\includegraphics{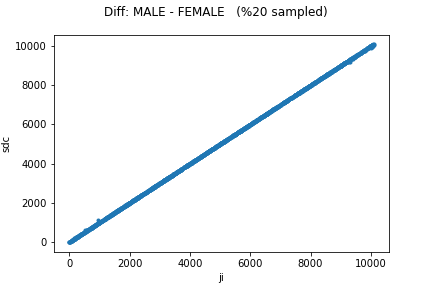}
\end{tabular}
\end{adjustbox}
\end{table}
\FloatBarrier

\begin{table}[htb]
\caption{Pointwise Mutual Information (pmi), Squared PMI (pmi\_2), Log-Likelihood Ratio (ll) Comparison} 
\begin{adjustbox}{width=1.1\textwidth,center}
\begin{tabular}{c@{}c}
\includegraphics{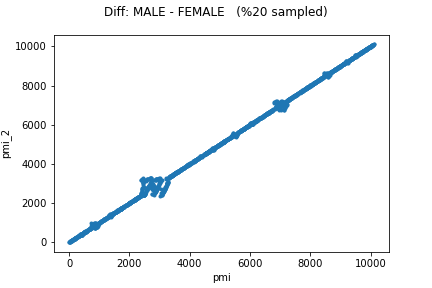} &
\includegraphics{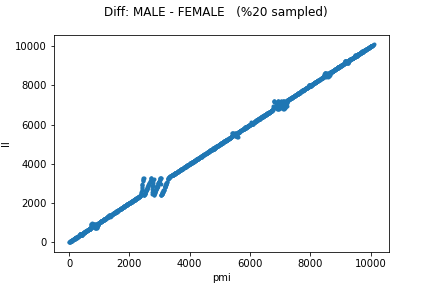} \\
\includegraphics{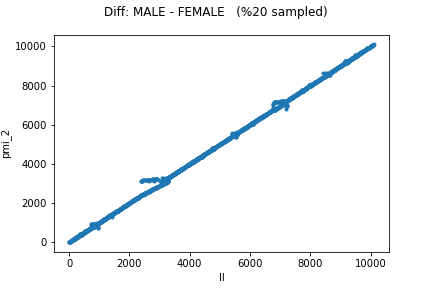}
\end{tabular}
\end{adjustbox}
\end{table}
\FloatBarrier

\begin{table}[htb]
\caption{Demographic Parity (di), Pointwise Mutual Information (pmi), Normalized Pointwise Mutual Information (npmi\_{xy}), $\tau_b$ (tau\_b) Comparison}
\begin{adjustbox}{width=\textwidth,center}
\begin{tabular}{cc}
\includegraphics{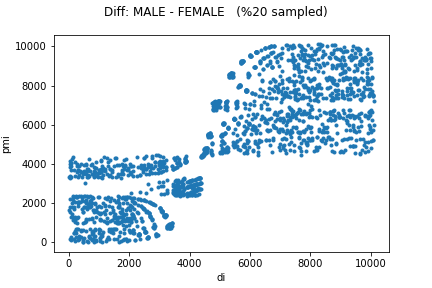} &
\includegraphics{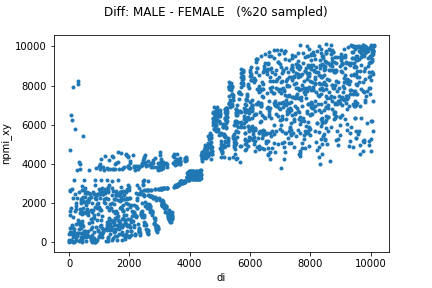} \\
\includegraphics{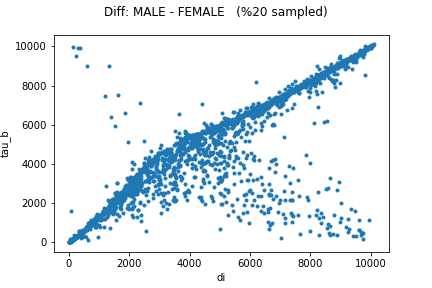} &
\includegraphics{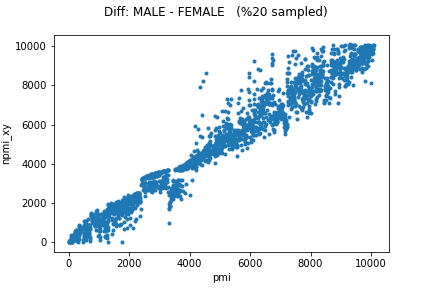} \\
\includegraphics{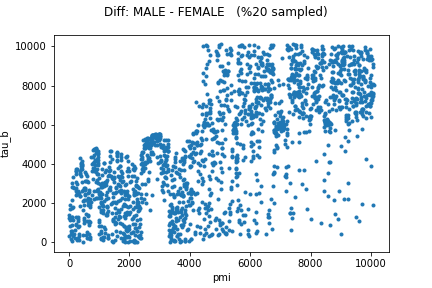} &
\includegraphics{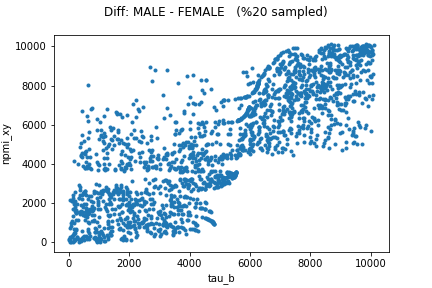}
\end{tabular}
\end{adjustbox}
\end{table}
\FloatBarrier

\subsection{Movement plots}\label{plot:movement}

\begin{table}[htb]
\caption{Demographic Parity (di), Jaccard Index (ji), Sørensen-Dice Coefficient (sdc) Comparison}
\begin{adjustbox}{width=1.25\linewidth,center}
\begin{tabular}{c}
\includegraphics{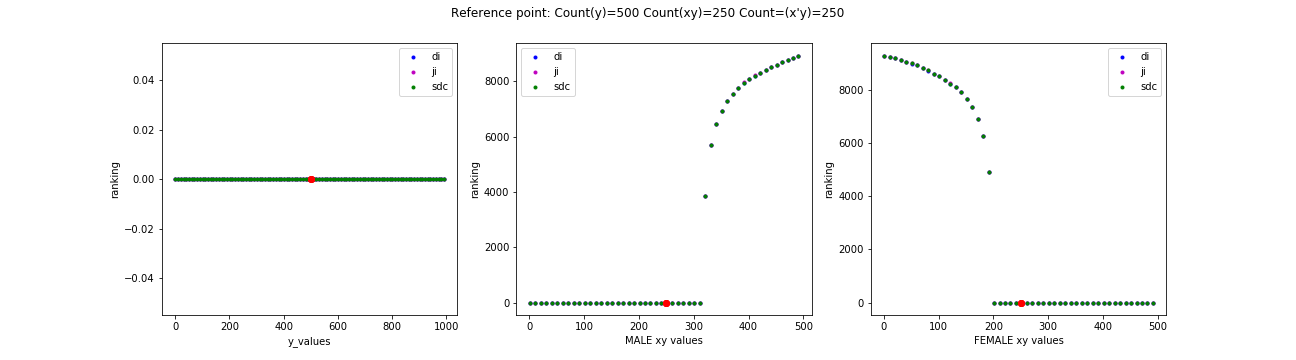} \\
\includegraphics{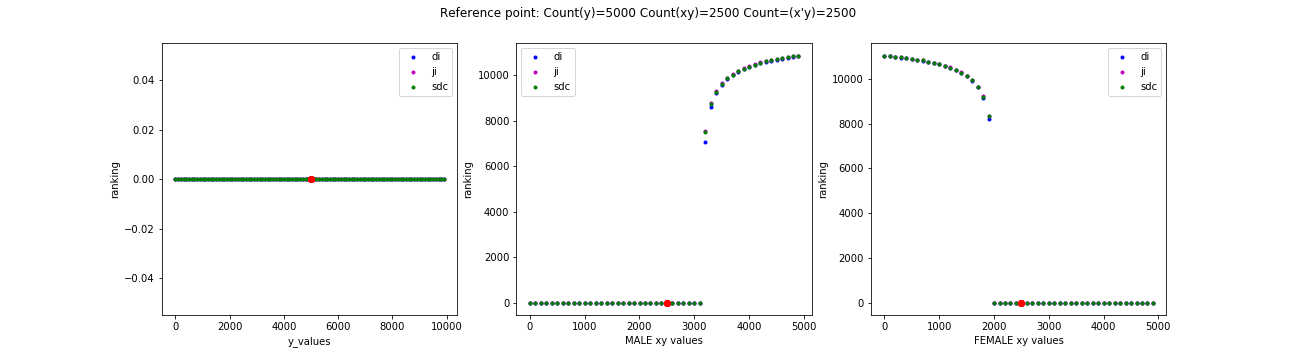} \\
\includegraphics{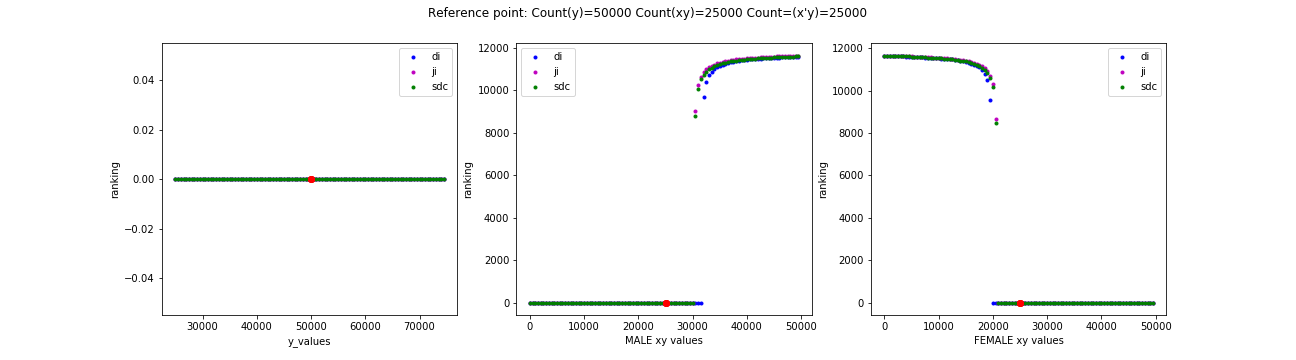}
\end{tabular}
\end{adjustbox}
\end{table}

\begin{table}[htb]
\caption{Pointwise Mutual Information (pmi), Squared PMI (pmi\_2), Log-Likelihood Ratio (ll) Comparison}
\begin{adjustbox}{width=1.25\linewidth,center}
\begin{tabular}{c}
\includegraphics{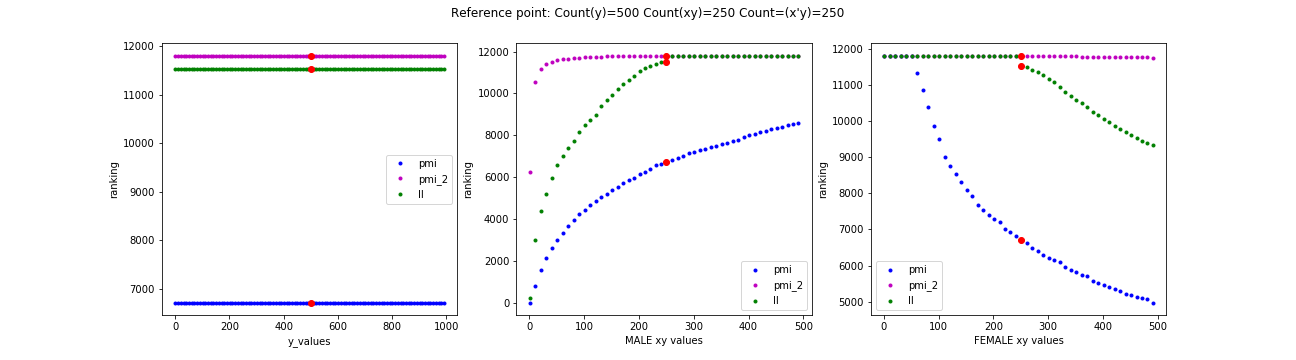} \\
\includegraphics{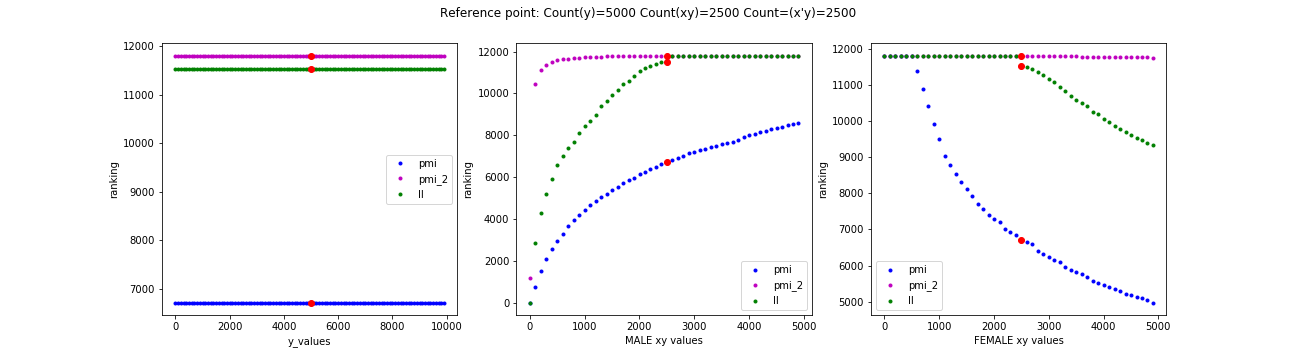} \\
\includegraphics{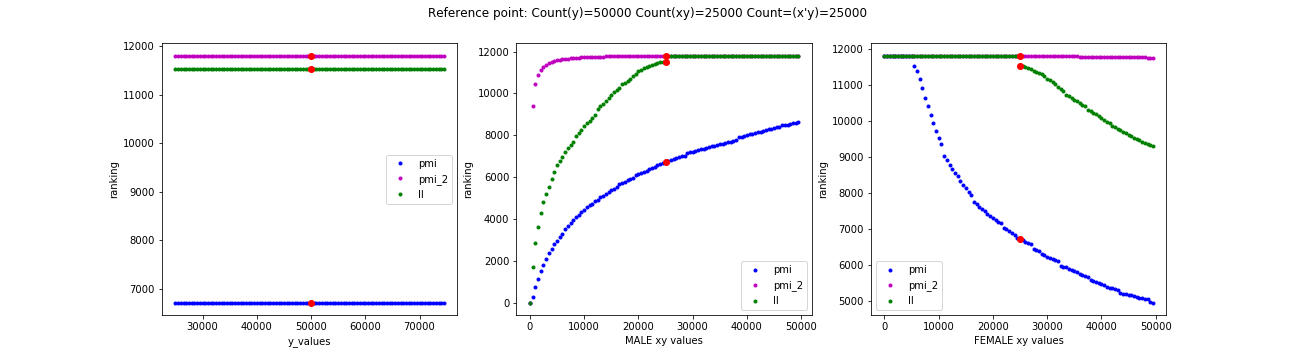}
\end{tabular}
\end{adjustbox}
\end{table}

\begin{table}[htb]
\caption{Demographic Parity (di), Pointwise Mutual Information (pmi), Normalized Pointwise Mutual Information (npmi\_{xy}), $\tau_b$ (tau\_b) Comparison}
\begin{adjustbox}{width=1.25\linewidth,center}
\begin{tabular}{l}
\includegraphics{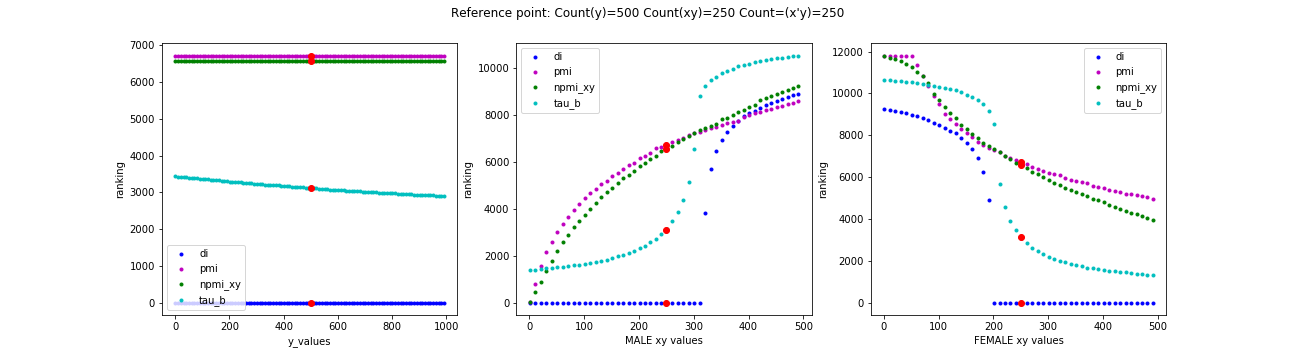} \\
\includegraphics{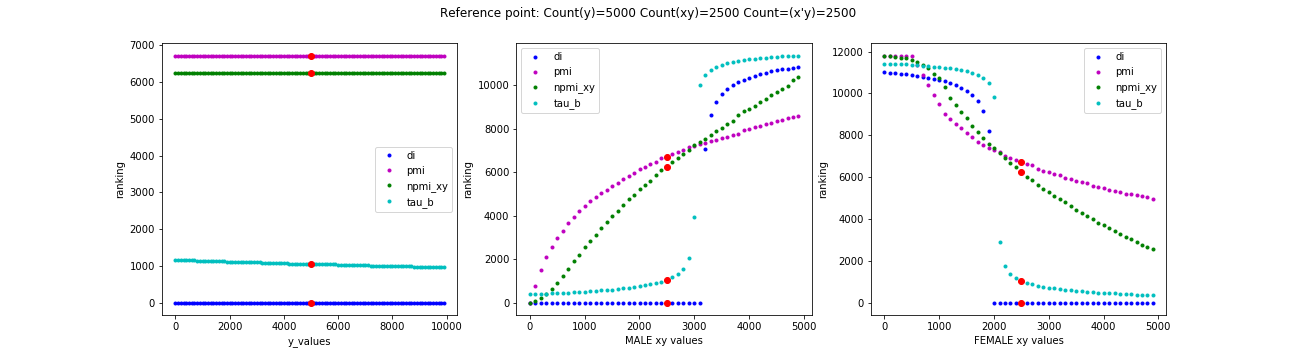} \\
\includegraphics{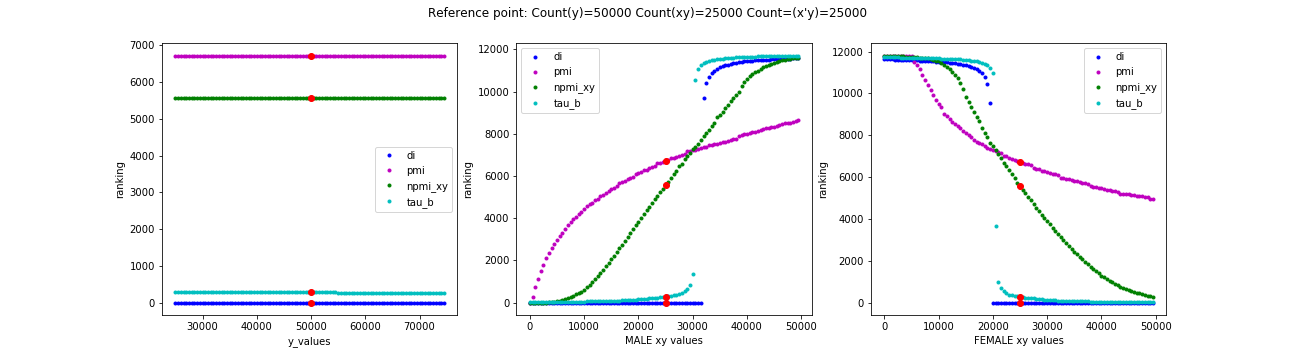}
\end{tabular}
\end{adjustbox}
\end{table}

\begin{table}
\caption{Example Top Results}
\begin{tabular}{ |c||c|c|c|c|c|c|  }
\hline
&\multicolumn{2}{|c|}{DP}&\multicolumn{2}{|c|}{PMI}&\multicolumn{2}{|c|}{nPMI$_{xy}$} \\
\hline
Ranks&Label&Count&Label&Count&Label&Count\\
\hline
0&Female&265853&Dido Flip&140&&610\\
1&Woman&270748&Webcam Model&184&Dido Flip&140\\
2&Girl&221017&Boho-chic&151&&2906\\
3&Lady&166186&&610&Eye Liner&3144\\
4&Beauty&562445&Treggings&126&Long Hair&56832\\
5&Long Hair&56832&Mascara&539&Mascara&539\\
6&Happiness&117562&&145&Lipstick&8688\\
7&Hairstyle&145151&Lace Wig&70&Step Cutting&6104\\
8&Smile&144694&Eyelash Extension&1167&Model&10551\\
9&Fashion&238100&Bohemian Style&460&Eye Shadow&1235\\
10&Fashion Designer&101854&&78&Photo Shoot&8775\\
11&Iris&120411&Gravure Idole&200&Eyelash Extension&1167\\
12&Skin&202360&&165&Boho-chic&460\\
13&Textile&231628&Eye Shadow&1235&Webcam Model&151\\
14&Adolescence&221940&&156&Bohemian Style&184\\
\hline
\end{tabular}
\end{table}

\end{document}


\maketitle

\tableofcontents{}

\section{Metric Alternatives}
\begin{itemize}
\item Disparate Impact
\[
DI\_gap = p(y|x_1) - p(y|x_2)
\]
\item Sørensen-Dice Coefficient
\[
SDC\_gap=
\frac{p(x_1,y)}{p(x_1)+p(y)}
-\frac{p(x_2,y)}{p(x_2)+p(y)}
\]
\item Jaccard Index
\[
JI\_gap=
\frac{p(x_1,y)}{p(x_1)+p(y)-p(x_1,y)}
-\frac{p(x_2,y)}{p(x_2)+p(y)-p(x_2,y)}
\]
\item Log-Likelihood Ratio of given y
\[
LL\_gap = ln(p(x_1|y)) - ln(p(x_2|y))
\]
\item PMI
\[
PMI\_gap=ln\left(\frac{p(x_1,y)}{p(x_1)p(y)}\right) -
ln\left(\frac{p(x_2,y)}{p(x_2)p(y)}\right)
\]
\item NPMI p(y) normalization
\[
NPMI_y\_gap=\frac
{ln\left(\frac{p(x_1,y)}{p(x_1)p(y)}\right)}
{ln\left(p(y)\right)} -
\frac
{ln\left(\frac{p(x_2,y)}{p(x_2)p(y)}\right)}
{ln\left(p(y)\right)}
\]
\item NPMI p(x,y) normalization
\[
NPMI_{xy}\_gap=\frac
{ln\left(\frac{p(x_1,y)}{p(x_1)p(y)}\right)}
{ln\left(p(x_1,y)\right)} -
\frac
{ln\left(\frac{p(x_2,y)}{p(x_2)p(y)}\right)}
{ln\left(p(x_2,y)\right)}
\]
\item PMI$^2$
\[
PMI^2\_gap=ln\left(\frac{p(x_1,y)^2}{p(x_1)p(y)}\right) -
ln\left(\frac{p(x_2,y)^2}{p(x_2)p(y)}\right)
\]
\item Kendall Rank Correlation($\tau_b$):
\[
\tau_b = \frac{n_c-n_d}{\sqrt{(n_0-n_1)(n_0-n_2)}}
\]
\[
\tau_b\_gap=\tau_b(l_1=x_1,l_2=y)-\tau_b(l_1=x_2,l_2=y)
\]
\item T-test:
\[
ttest\_gap=\frac{p(x_1,y)-p(x_1)*p(y)}{\sqrt{p(x_1)*p(y)}}
- \frac{p(x_2,y)-p(x_2)*p(y)}{\sqrt{p(x_2)*p(y)}}
\]
\end{itemize}

\section{Mathematical Reductions}
\subsection{Overview}
\[
DI\_gap = p(y|x_1) - p(y|x_2)
\]

\[
SDC\_gap=\frac{p(x_1,y)}{p(x_1)+p(y)}-\frac{p(x_2,y)}{p(x_2)+p(y)}
\]

\[
LL\_gap = ln(p(x_1|y)) - ln(p(x_2|y))
\]

\[
PMI\_gap=
ln(p(y|x_1))-ln(p(y|x_2))
\]

\[
PMI^2\_gap=ln(p(y|x_1))-ln(p(y|x_2))+\frac{ln(p(x_1,y))}{ln(p(x_2,y))}
\]

\[
NPMI_y\_gap=
\frac{ln(p(y|x_1))}{ln(p(y))}
-\frac{ln(p(y|x_2))}{ln(p(y))}
\]

\[
NPMI_{xy}\_gap=
\frac{ln(p(y|x_1))}{ln(p(x_1,y))}
- \frac{ln(p(y|x_2))}{ln(p(x_2,y))}
+ \frac{ln(p(y))}{ln(p(x_1,y))ln(p(x_2,y))}
. ln(\frac{p(x_2,y)}{p(x_1,y)})
\]

\[
\tau_b\_gap=\tau_b(l_1=x_1,l_2=y)-\tau_b(l_1=x_2,l_2=y)
\]
\[
ttest\_gap=\frac{p(x_1,y)-p(x_1)*p(y)}{\sqrt{p(x_1)*p(y)}}
- \frac{p(x_2,y)-p(x_2)*p(y)}{\sqrt{p(x_2)*p(y)}}
\]

\subsection{Reduction Steps}
\textbf{PMI:}
\[
PMI\_gap=ln\left(\frac{p(x_1,y)}{p(x_1)p(y)}\right) -
ln\left(\frac{p(x_2,y)}{p(x_2)p(y)}\right)
\]

\[
PMI\_gap=ln\left(
\frac{p(x_1,y)}{p(x_1)p(y)}.
\frac{p(x_2)p(y)}{p(x_2,y)}\right)
\]

\[
PMI\_gap=ln\left(
\frac{p(x_1,y)}{p(x_1)}.
\frac{p(x_2)}{p(x_2,y)}\right)
\]

\[
PMI\_gap=ln\left(
\frac{p(y|x_1)}{p(y|x_2)}\right)
\]

\[
PMI\_gap=
ln(p(y|x_1))-ln(p(y|x_2))
\]
\textbf{NPMI p(y) normalization:}
\[
NPMI\_gap=\frac
{ln\left(\frac{p(x_1,y)}{p(x_1)p(y)}\right)}
{ln\left(p(y)\right)} -
\frac
{ln\left(\frac{p(x_2,y)}{p(x_2)p(y)}\right)}
{ln\left(p(y)\right)}
\]

\[
NPMI\_gap=
\frac{ln(p(y|x_1))-ln(p(y|x_2))}{ln(p(y))}
\]
\textbf{NPMI p(x,y) normalization:}
\[
NPMI\_gap=\frac
{ln\left(\frac{p(x_1,y)}{p(x_1)p(y)}\right)}
{ln\left(p(x_1,y)\right)} -
\frac
{ln\left(\frac{p(x_2,y)}{p(x_2)p(y)}\right)}
{ln\left(p(x_2,y)\right)}
\]

\[
NPMI\_gap=
ln\left({\left(\frac{p(x_1,y)}{p(x_1)p(y)}\right)}^{1/ln(p(x_1,y))}\right) -
ln\left({\left(\frac{p(x_2,y)}{p(x_2)p(y)}\right)}^{1/ln(p(x_2,y))}\right)
\]

\[
NPMI\_gap=
ln\left(
\frac
{{\left(\frac{p(x_1,y)}{p(x_1)p(y)}\right)}^{1/ln(p(x_1,y))}}
{{\left(\frac{p(x_2,y)}{p(x_2)p(y)}\right)^{1/ln(p(x_2,y))}}}
\right)
\]

\[
NPMI\_gap=
ln\left(
\frac
{{\left(\frac{p(x_1,y)}{p(x_1)}\right)}^{1/ln(p(x_1,y))}}
{{\left(\frac{p(x_2,y)}{p(x_2)}\right)^{1/ln(p(x_2,y))}}}
.{p(y)}^{(1/ln(p(x_1,y))-1/ln(p(x_2,y)))}
\right)
\]

\[
NPMI\_gap=
ln\left(
\frac
{{\left(\frac{p(x_1,y)}{p(x_1)}\right)}^{1/ln(p(x_1,y))}}
{{\left(\frac{p(x_2,y)}{p(x_2)}\right)^{1/ln(p(x_2,y))}}}
\right)
+ln\left(
{p(y)}^{(1/ln(p(x_1,y))-1/ln(p(x_2,y)))}
\right)
\]

\[
NPMI\_gap=
ln\left(
\frac
{{p(y|x_1)}^{1/ln(p(x_1,y))}}
{{p(y|x_2)^{1/ln(p(x_2,y))}}}
\right)
+ ln(p(y)).\left(\frac{1}{ln(p(x_1,y))}-\frac{1}{ln(p(x_2,y))}\right)
\]

\[
NPMI\_gap=
\frac{ln(p(y|x_1))}{ln(p(x_1,y))}
- \frac{ln(p(y|x_2))}{ln(p(x_2,y))}
+ ln(p(y)).\left(\frac{ln(\frac{p(x_2,y)}{p(x_1,y)})}{ln(p(x_1,y))ln(p(x_2,y))}\right)
\]

\[
NPMI\_gap=
\frac{ln(p(y|x_1))}{ln(p(x_1,y))}
- \frac{ln(p(y|x_2))}{ln(p(x_2,y))}
+ \frac{ln(p(y))ln(\frac{p(x_2,y)}{p(x_1,y)})}{ln(p(x_1,y))ln(p(x_2,y))}
\]
\textbf{PMI$^2$:}
\[
PMI^2\_gap=ln\left(\frac{p(x_1,y)^2}{p(x_1)p(y)}\right) -
ln\left(\frac{p(x_2,y)^2}{p(x_2)p(y)}\right)
\]

\[
PMI^2\_gap=ln\left(
\frac{p(x_1,y)^2}{p(x_1)p(y)}.
\frac{p(x_2)p(y)}{p(x_2,y)^2}\right)
\]

\[
PMI^2\_gap=ln\left(
\frac{p(x_1,y)^2}{p(x_1)}.
\frac{p(x_2)}{p(x_2,y)^2}\right)
\]

\[
PMI^2\_gap=ln\left(
\frac{p(y|x_1)p(x_1,y)}{p(y|x_2)p(x_2,y)}\right)
\]

\[
PMI^2\_gap=ln(p(y|x_1))-ln(p(y|x_2))+ln(p(x_1,y))-ln(p(x_2,y))
\]
\textbf{Kendall Rank Correlation($\tau_b$):} \\

Formal definition:
\[
\tau_b = \frac{n_c-n_d}{\sqrt{(n_0-n_1)(n_0-n_2)}}
\]
\begin{itemize}
    \item $n_0=n(n-1)/2$
    \item $n_1=\sum_{i}t_i(t_i-1)/2$
    \item $n_2=\sum_{j}u_j(u_j-1)/2$
    \item $n_c$ is number of concordant pairs
    \item $n_d$ is number of discordant pairs
    \item $n_c$ is number of concordant pairs
    \item $t_i$ is number of tied values in the $i^{th}$ group of ties for the first quantity
    \item $u_j$ is number of tied values in the $j^{th}$ group of ties for the second quantity
\end{itemize}

New notations for our use case:
\[
\tau_b(l_1=x_1,l_2=y) = \frac{n_c(l_1=x_1,l_2=y)-n_d(l_1=x_1,l_2=y)}{\sqrt{({n \choose 2}-n_s(l=x_1))({n\choose 2}-n_s(l=y))}}
\]
\begin{itemize}
    \item $n_c(l_1,l_2)={C_{l_1=0,l_2=0}\choose 2}+{C_{l_1=1,l_2=1}\choose 2}$
    \item $n_d(l_1,l_2)={C_{l_1=0,l_2=1}\choose 2}+{C_{l_1=1,l_2=0}\choose 2}$
    \item $n_s(l)={C_{l=0}\choose 2}+{C_{l=1}\choose 2}$
    \item $C_{conditions}$ is number of examples which satisfies the conditions
    \item $n$ number of examples in data set
\end{itemize}

Gap is defined as similarly:
\[
\tau_b\_gap=\tau_b(l_1=x_1,l_2=y)-\tau_b(l_1=x_2,l_2=y)
\]

\section{Metric Orientations}
\subsection{Overview}
\begin{tabular}{ |c|c|c|c|  }
\hline
&$\partial p(y)$&$\partial p(x_1, y)$&$\partial p(x_2, y)$\\
\hline
&&&\\[0.2ex]
$\partial$DI&$0$&$\frac{1}{p(x_1)}$&$\frac{-1}{p(x_2)}$\\[2.5ex]
$\partial$PMI&$0$&$\frac{p(x_1)}{p(x_1,y)p(y)}$&$\frac{-p(x_2)}{p(x_2,y)p(y)}$\\[2.5ex]
$\partial nPMI_{y}$&$\frac{ln(\frac{p(x_2,y)}{p(x_1,y)})}{ln^2(p(y))p(y)}$&$\frac{1}{ln(p(y))p(x_1, y)}$&$\frac{-1}{ln(p(y))p(x_2, y)}$\\[2.5ex]
$\partial nPMI_{xy}$&$\frac{ln(\frac{p(x_2,y)}{p(x_1,y)})}{ln(p(x_2,y))ln(p(x_1,y))p(y)}$&\textit{*check App-3.2}&\textit{*check App-3.2}\\[2.5ex]
$\partial PMI^2$&0&$\frac{p(x_1)}{p(x_1,y)p(y)}+\frac{1}{ln(p(x_2,y))p(x_1,y)}$&$\frac{-p(x_2)}{p(x_2,y)p(y)}+\frac{-ln(p(x_1,y))}{ln(p(x_2,y))^2p(x_2,y)}$\\[2.5ex]
$\partial$SDC&$\frac{p(x_1,y)}{(p(x_1)+p(y))^2}-\frac{p(x_2,y)}{(p(x_2)+p(y))^2}$&$\frac{1}{(p(x_1)+p(y))^2}$&$\frac{-1}{(p(x_2)+p(y))^2}$\\[2.5ex]
$\partial$JI&\textit{*check App-3.2}&$\frac{p(x_1)+p(y)}{(p(x_1)+p(y)-p(x_1,y))^2}$&$\frac{p(x_2)+p(y)}{(p(x_2)+p(y)-p(x_2,y))^2}$\\[2.5ex]
$\partial$LLR&$0$&$\frac{1}{p(x_1, y)}$&$\frac{-1}{p(x_2, y)}$\\[2.5ex]
$\partial \tau_b$&\textit{*check App-3.2}&\textit{*check App-3.2}&\textit{*check App-3.2}\\[2.5ex]
$\partial ttest\_gap$&\textit{*check App-3.2}&\textit{*check App-3.2}&\textit{*check App-3.2}\\[2.5ex]
\hline
\end{tabular}

\subsection{Orientation Details}
\textbf{Disparate Impact:}
\[
DI\_gap = p(y|x_1) - p(y|x_2)
\]
\[
\frac{\partial DI\_gap}{\partial p(y)} = 0
\]
\[
\frac{\partial DI\_gap}{\partial p(x_1, y)} = \frac{1}{p(x_1)}
\]
\[
\frac{\partial DI\_gap}{\partial p(x_2, y)} = \frac{-1}{p(x_2)}
\]
\textbf{Sørensen-Dice Coefficient:}
\[
SDC\_gap=\frac{p(x_1,y)}{p(x_1)+p(y)}-\frac{p(x_2,y)}{p(x_2)+p(y)}
\]
\[
\frac{\partial SDC\_gap}{\partial p(y)}=\frac{p(x_1,y)}{(p(x_1)+p(y))^2}-\frac{p(x_2,y)}{(p(x_2)+p(y))^2}
\]
\[
\frac{\partial SDC\_gap}{\partial p(x_1, y)}=\frac{1}{(p(x_1)+p(y))^2}
\]
\[
\frac{\partial SDC\_gap}{\partial p(x_2, y)}=\frac{-1}{(p(x_2)+p(y))^2}
\]
\textbf{Jaccard Index}
\[
JI\_gap=
\frac{p(x_1,y)}{p(x_1)+p(y)-p(x_1,y)}
-\frac{p(x_2,y)}{p(x_2)+p(y)-p(x_2,y)}
\]
\[
\frac{\partial JI\_gap}{\partial p(y)}=
\frac{p(x_1,y)}{(p(x_1)+p(y)-p(x_1,y))^2}
-\frac{p(x_2,y)}{(p(x_2)+p(y)-p(x_2,y))^2}
\]
\[
\frac{\partial JI\_gap}{\partial p(x_1,y)}=
\frac{p(x_1)+p(y)}{(p(x_1)+p(y)-p(x_1,y))^2}
\]
\[
\frac{\partial JI\_gap}{\partial p(x_2,y)}=
\frac{p(x_2)+p(y)}{(p(x_2)+p(y)-p(x_2,y))^2}
\]
\textbf{Log-Likelihood Ratio:}
\[
LL\_gap = ln(p(x_1|y)) - ln(p(x_2|y))
\]
\[
\frac{\partial LL\_gap}{\partial p(y)} = \frac{p(x_1,y)}{p(x_1|y)p(y)^2} - \frac{p(x_2,y)}{p(x_2|y)p(y)^2} = 0
\]
\[
\frac{\partial LL\_gap}{\partial p(x_1, y)} = \frac{1}{p(x_1|y)p(y)} = \frac{1}{p(x_1, y)}
\]
\[
\frac{\partial LL\_gap}{\partial p(x_2, y)} = \frac{-1}{p(x_2, y)}
\]
\textbf{PMI:}
\[
PMI\_gap=
ln(p(y|x_1))-ln(p(y|x_2))
\]
\[
\frac{\partial PMI\_gap}{\partial p(y)} = 0
\]
\[
\frac{\partial PMI\_gap}{\partial p(x_1, y)} = \frac{p(x_1)}{p(x_1,y)p(y)}
\]
\[
\frac{\partial PMI\_gap}{\partial p(x_1, y)} = \frac{-p(x_2)}{p(x_2,y)p(y)}
\]
\textbf{PMI$^2$:}
\[
PMI^2\_gap=ln(p(y|x_1))-ln(p(y|x_2))+\frac{ln(p(x_1,y))}{ln(p(x_2,y))}
\]
\[
\frac{\partial PMI^2\_gap}{\partial p(y)} = 0
\]
\[
\frac{\partial PMI^2\_gap}{\partial p(x_1, y)} = \frac{p(x_1)}{p(x_1,y)p(y)} + \frac{1}{ln(p(x_2,y))p(x_1,y)}
\]
\[
\frac{\partial PMI^2\_gap}{\partial p(x_2, y)} = \frac{-p(x_2)}{p(x_2,y)p(y)} + \frac{-ln(p(x_1,y))}{ln(p(x_2,y))^2p(x_2,y)}
\]
\textbf{NPMI, normalized by p(y) :}
\[
NPMI\_gap = \frac{ln(p(y|x_1))}{ln(p(y))} - \frac{ln(p(y|x_2))}{ln(p(y))}
\]
\[
\frac{\partial NPMI\_gap}{\partial p(y)} = \frac{ln(p(y|x_2))}{ln^2(p(y))p(y)} - \frac{ln(p(y|x_1))}{ln^2(p(y))p(y)} = \frac{ln(\frac{p(x_2,y)}{p(x_1,y)})}{ln^2(p(y))p(y)}
\]
\[
\frac{\partial NPMI\_gap}{\partial p(x_1,y)} = \frac{1}{ln(p(y))p(x_1, y)}
\]
\[
\frac{\partial NPMI\_gap}{\partial p(x_2,y)} = \frac{-1}{ln(p(y))p(x_2, y)}
\]
\textbf{NPMI, normalized by p(x,y) :}\\
\[
NPMI\_gap=
\frac{ln(p(y|x_1))}{ln(p(x_1,y))}
- \frac{ln(p(y|x_2))}{ln(p(x_2,y))}
+ \frac{ln(p(y))}{ln(p(x_1,y))ln(p(x_2,y))}
. ln(\frac{p(x_2,y)}{p(x_1,y)})
\]
\textbf{Kendall rank correlation($\tau_b$):}\\
\[
\tau_b\_gap=\tau_b(l_1=x_1,l_2=y)-\tau_b(l_1=x_2,l_2=y)
\]
\textbf{T-test:}\\
\[
ttest\_gap=\frac{p(x_1,y)-p(x_1)*p(y)}{\sqrt{p(x_1)*p(y)}}
- \frac{p(x_2,y)-p(x_2)*p(y)}{\sqrt{p(x_2)*p(y)}}
\]

\section{Comparison Tables}
\subsection{Mean/Variance of top100 Male}
\begin{tabular}{ |p{2cm}||p{3.5cm}|p{3.5cm}|  }
\hline
\multicolumn{3}{|c|}{Scales of top100 ($x_1$=MALE)} \\
\hline
Metrics&Mean/Std $C(y)$&Mean/Std $C(x_1,y)$\\
\hline
di&$101.38(\pm199.32)$&$1.00(\pm0.00)$\\
npmi\_xy&$24371.15(\pm27597.10)$&$237.92(\pm604.29)$\\
npmi\_y&$75160.45(\pm119779.13)$&$5680.44(\pm17193.19)$\\
pmi&$1875.86(\pm3344.23)$&$2.43(\pm4.72)$\\
pmi\_2&$831.43(\pm620.02)$&$1.09(\pm0.32)$\\
sdc&$676.33(\pm412.23)$&$1.00(\pm0.00)$\\
tau\_b&$189294.69(\pm149505.10)$&$24524.89(\pm32673.77)$\\
\hline
\end{tabular}

\subsection{Mean/Variance of top100 Male-Female Gap}

\begin{table}
\begin{tabular}{|p{2cm}||l|l|l|}
\hline
\multicolumn{4}{|c|}{Counts for top100 gaps($x_1$=MALE, $x_2$=FEMALE)} \\
\hline
Metrics&Mean/Std $C(y)$&Mean/Std $C(x_1,y)$&Mean/Std $C(x_2, y)$\\
\hline
di&154960.81($\pm$147823.81)& 71552.62($\pm$67400.44)& 70398.07($\pm$56824.32)\\
ji& 70403.46($\pm$86766.14)& 29670.94($\pm$35381.09)& 35723.13($\pm$39308.46)\\
ll&  1020.60($\pm$1971.77)&    57.95($\pm$178.59)&   558.36($\pm$1420.15)\\
npmi\_xy& 13869.42($\pm$46244.17)&  5771.73($\pm$23564.63)&  9664.46($\pm$31475.19)\\
npmi\_y& 23156.11($\pm$73662.90)&  6632.79($\pm$25631.26)& 10376.39($\pm$33559.27)\\
pmi&  1020.60($\pm$1971.77)&    57.95($\pm$178.59)&   558.36($\pm$1420.15)\\
pmi\_2&  1020.60($\pm$1971.77)&    57.95($\pm$178.59)&   558.36($\pm$1420.15)\\
sdc& 67648.20($\pm$87258.41)& 28193.83($\pm$35295.25)& 34443.22($\pm$39501.21)\\
tau\_b&134076.65($\pm$144354.70)& 53845.80($\pm$58016.12)& 56032.09($\pm$52124.92)\\
\hline
\end{tabular}
\end{table}

\subsection{Minimum/Maximum of top100 Male-Female Gap}

\begin{tabular}{ |c|c|c|c|  }
\hline
Metrics&Min/Max $C(y)$&Min/Max $C(x_1,y)$&Min/Max $C(x_2, y)$\\
\hline
DI&     6104/785045   &      628/239950   &     5347/197795   \\
PMI&       15/10551    &        1/1059     &        8/7755     \\
$nPMI_{y}$&       35/562445   &        1/144185   &        9/183132   \\
$nPMI_{xy}$&       34/270748   &        1/144185   &       20/183132   \\
$PMI^2$&       15/10551    &        1/1059     &        8/7755     \\
SDC&     2906/562445   &      139/144185   &     2563/183132   \\
JI&     4158/562445   &      399/144185   &     3359/183132   \\
LL&       15/10551    &        1/1059     &        8/7755     \\
$\tau_b$&     6104/785045   &      628/207723   &     5347/183132   \\
T-test&      960/562445   &       72/144185   &      870/183132   \\
\hline
\end{tabular}

\section{Comparison Plots}
\subsection{Overall Rank Changes}
\subsubsection{DI, JI, SDC Comparison}
\includegraphics[width=5.8cm]{images/appendix_images/di_VS_ji.png}
\includegraphics[width=5.8cm]{images/appendix_images/di_VS_sdc.png}
\includegraphics[width=5.8cm]{images/appendix_images/ji_VS_sdc.png}

\subsubsection{PMI, PMI$^2$, LLR Comparison}
\includegraphics[width=5.8cm]{images/appendix_images/pmi_VS_pmi_2.png}
\includegraphics[width=5.8cm]{images/appendix_images/pmi_VS_ll.png}
\includegraphics[width=5.8cm]{images/appendix_images/ll_VS_pmi_2.png}

\subsubsection{DI, PMI, NPMI$_{xy}$, $\tau_b$ Comparison}
\includegraphics[width=5.8cm]{images/appendix_images/di_VS_pmi.png}
\includegraphics[width=5.8cm]{images/appendix_images/di_VS_npmi_xy.png}
\includegraphics[width=5.8cm]{images/appendix_images/di_VS_tau_b.png}
\includegraphics[width=5.8cm]{images/appendix_images/pmi_VS_npmi_xy.png}
\includegraphics[width=5.8cm]{images/appendix_images/pmi_VS_tau_b.png}
\includegraphics[width=5.8cm]{images/appendix_images/tau_b_VS_npmi_xy.png}

\subsection{Movement plots}
\subsubsection{DI, JI, SDC Comparison}

\includegraphics[height=5.5cm]{images/appendix_images/small_di_ji_sdc.png}
\includegraphics[height=5.5cm]{images/appendix_images/medium_di_ji_sdc.png}
\includegraphics[height=5.5cm]{images/appendix_images/large_di_ji_sdc.png}

\subsubsection{PMI, PMI$^2$, LLR Comparison}

\includegraphics[height=5.5cm]{images/appendix_images/small_pmi_pmi_2_ll.png}
\includegraphics[height=5.5cm]{images/appendix_images/medium_pmi_pmi_2_ll.png}
\includegraphics[height=5.5cm]{images/appendix_images/large_pmi_pmi_2_ll.png}

\subsubsection{DI, PMI, NPMI$_{xy}$, $\tau_b$ Comparison}

\includegraphics[height=5.5cm]{images/appendix_images/small_di_pmi_npmi_xy_tau_b.png}
\includegraphics[height=5.5cm]{images/appendix_images/medium_di_pmi_npmi_xy_tau_b.png}
\includegraphics[height=5.5cm]{images/appendix_images/large_di_pmi_npmi_xy_tau_b.png}

\section{Example Top Result}

\begin{tabular}{ |c||c|c|c|c|c|c|  }
\hline
&\multicolumn{2}{|c|}{DP}&\multicolumn{2}{|c|}{PMI}&\multicolumn{2}{|c|}{NPMI$_{xy}$} \\
\hline
Ranks&Label&Count&Label&Count&Label&Count\\
\hline
0&Female&265853&Dido Flip&140&&610\\
1&Woman&270748&Webcam Model&184&Dido Flip&140\\
2&Girl&221017&Boho-chic&151&&2906\\
3&Lady&166186&&610&Eye Liner&3144\\
4&Beauty&562445&Treggings&126&Long Hair&56832\\
5&Long Hair&56832&Mascara&539&Mascara&539\\
6&Happiness&117562&&145&Lipstick&8688\\
7&Hairstyle&145151&Lace Wig&70&Step Cutting&6104\\
8&Smile&144694&Eyelash Extension&1167&Model&10551\\
9&Fashion&238100&Bohemian Style&460&Eye Shadow&1235\\
10&Fashion Designer&101854&&78&Photo Shoot&8775\\
11&Iris&120411&Gravure Idole&200&Eyelash Extension&1167\\
12&Skin&202360&&165&Boho-chic&460\\
13&Textile&231628&Eye Shadow&1235&Webcam Model&151\\
14&Adolescence&221940&&156&Bohemian Style&184\\
\hline
\end{tabular}